\documentclass[a4paper,fleqn]{cas-sc}

\usepackage[authoryear,longnamesfirst]{natbib}
\usepackage{framed,multirow}

\usepackage{amssymb}
\usepackage{latexsym}

\usepackage{url}
\usepackage{xcolor}

\usepackage{times}
\usepackage{epsfig}
\usepackage{graphicx}
\usepackage{amsmath}
\usepackage{float}
\usepackage{subfigure}
\usepackage{standalone}
\usepackage{paralist}

\usepackage{lineno,hyperref} 

\usepackage{amssymb}
\usepackage{amsbsy}

\usepackage{import}
\usepackage{pgfplots}
\usepackage{tikz}
\usepackage{tpic}
\usepackage{adjustbox}
\usepackage{pgfplotstable}

\usepackage{grffile}
\usetikzlibrary{arrows,shapes,chains}
\usetikzlibrary{plotmarks}
\usetikzlibrary{arrows.meta}
\pgfplotsset{compat=1.13}

\usetikzlibrary{backgrounds,arrows}
\usetikzlibrary{automata, decorations.pathmorphing, positioning}
\usetikzlibrary{shapes.geometric,shapes.multipart}
\usetikzlibrary{fadings}

\def\tsc#1{\csdef{#1}{\textsc{\lowercase{#1}}\xspace}}
\tsc{WGM}
\tsc{QE}
\tsc{EP}
\tsc{PMS}
\tsc{BEC}
\tsc{DE}

\begin{document}
	\let\WriteBookmarks\relax
	\def\floatpagepagefraction{1}
	\def\textpagefraction{.001}
	\shorttitle{MKF for Nonstationary Noise: An extension of BF using image context}
	\shortauthors{Feihong Liu et~al.}
	
	\title [mode = title]{Multi-Kernel Filtering for Nonstationary Noise: An Extension of Bilateral Filtering Using Image Context}                      
	\tnotemark[1]
	
	\tnotetext[1]{This work was supported in part by the National Key Research and Development Program of China under grant (2017YFB1002504), and by NIH grants (NS093842 and EB022880).}
	

	\author[1,3]{Feihong Liu}[ type=editor, 
	auid=000,bioid=1, 
	orcid=0000-0001-5199-5261]
	\ead[url]{child@stumail.nwu.edu.cn}

	\author[1,2]{Jun Feng}[auid=001,bioid=2,orcid=0000-0002-0706-2103]
	\cormark[1]
	\ead[URL]{fengjun@nwu.edu.cn}
	
	\author[3]{Pew-Thian Yap}[auid=002,bioid=3, orcid=0000-0003-1489-2102] 
	\cormark[1]
	\ead[URL]{ptyap@med.unc.edu}

	\author[3,4]{Dinggang Shen}[auid=003,bioid=4]
	\cormark[1]
	\ead[URL]{dgshen@med.unc.edu}
	
	\cortext[cor1]{Corresponding authors}
	
	\address[1]{School of Information Science and Technology, Northwest Universtiy, Xi'an, China}
	\address[2]{State-Province Joint Engineering and Research Center of Advanced Networking and Intelligent Information Services, School of Information Science and Technology, Northwest University, Xi'an, China}
	\address[3]{Department of Radiology and Biomedical Research Imaging Center (BRIC), University of North Carolina at Chapel Hill, NC, U.S.A.}
	\address[4]{Department of Brain and Cognitive Engineering, Korea University, Seoul, Korea}

	
	\nonumnote{The authors have declared that no competing interests exist. }
	
	\begin{abstract}
		Bilateral filtering (BF) is one of the most classical denoising filters, however, the manually initialized filtering kernel hampers its adaptivity across images with various characteristics. To deal with image variation (i.e., non-stationary noise), in this paper, we propose multi-kernel filter (MKF) which adapts filtering kernels to specific image characteristics automatically. The design of MKF takes inspiration from adaptive mechanisms of human vision that make full use of information in a visual context. More specifically, for simulating the visual context and its adaptive function, we construct the image context based on which we simulate the contextual impact on filtering kernels. 
		We first design a hierarchically clustering algorithm to generate a hierarchy of large to small coherent image patches, organized as a cluster tree, so that obtain multi-scale image representation. The leaf cluster and corresponding predecessor clusters are used to generate one of multiple range kernels that are capable of catering to image variation. 
		At first, we design a hierarchically clustering framework to generate a hierarchy of large to small coherent image patches that organized as a cluster tree, so that obtain multi-scale image representation, i.e., the image context. 
		Next, a leaf cluster is used to generate one of the multiple kernels, and two corresponding predecessor clusters are used to fine-tune the adopted kernel. Ultimately, the single spatially-invariant kernel in BF becomes multiple spatially-varying ones. We evaluate MKF on two public datasets, BSD$300$ and BrainWeb which are added integrally-varying noise and spatially-varying noise, respectively. Extensive experiments show that MKF outperforms state-of-the-art filters w.r.t.\,both mean absolute error and structural similarity. 		 
	\end{abstract}
	

	\maketitle

\section{Introduction}
\label{sec:S1}

Image filtering is a crucial preprocessing approach, which serves in a number of practical tasks, including noise removal~\citep{veraart2016denoising}, edge detection~\citep{wei2018joint}, and phase correction~\citep{eichner2015real}. Those filters aim to restore noise-free image from a noisy observation, 
\begin{equation}
\label{eq:e1}
I(\vec{x}) = O(I(\vec{x})) + n(\vec{x})  
\end{equation}
\noindent where the observed image $I(\vec{x})$ can be split into two channels, $O(I(\vec{x}))$, which is the denoised image, and $n(\vec{x})$, which is the residual noise. Conventional filters assume $I(\vec{x})$ with characteristics: 
\begin{inparaenum}[(1)]
	\item the noise distribution is stationary; and 
	\item the intensity gradient smaller than noise level is treated as noise.  
\end{inparaenum}
However, both assumptions are not always true. The noise level could be changed integrally and regionally over spatial, namely \textit{integrally-varying noise} and \textit{spatially-varying noise}. Both types of noise degenerate conventional filters significantly. 

While noise variabilities break above-mentioned assumptions, the human visual system deals with stimuli variation across circumstances effectively. A recent study unveils that a visual context adapts low-level features and makes them suit the context involuntarily, by which the same stimulus may lead to different visual  awareness~\citep{bar2004visual}. Lu et al. ($2018$) query whether the feature invariance is always an optimal strategy for biological or artificial vision~\citet{lu2018revealing}. Theoretically, the visual context comprises associated objects which may be formed in visual working memory~\citep{gao2011perceptual, eriksson2015neurocognitive, thiele2018neuromodulation}, where follows Gestalt grouping rules~\citep{peterson2013gestalt, gao2016organization}. On the other hand, because these Gestalt grouping rules have inspired computer vision researchers to design clustering algorithms~\citep{shi2000normalized, liu2017normalized}, we can generate image context using clustering to simulate visual context. And next, we exploit image context to affect low-level features extracted from an image.

In this paper, we propose multi-kernel filter (MKF) for improving adaptivity of bilateral filtering (BF). We first design a hierarchically clustering framework, following Gestalt grouping rules and top-down fashion so that generate a hierarchy of large to small coherent clusters (Sec.~\ref{subsec:s311}). The adopted clusters are organized as a cluster tree for constructing image context (Sec.~\ref{subsec:s312}). A leaf cluster of the tree is used to generate one of multiple filtering kernels; these kernels are next fine-tuned by corresponding predecessor clusters in higher levels of the hierarchy (Sec.~\ref{subsec:s32}). With our extension, the single spatially-invariant range kernel in BF, which is manually initialized, now, becomes multiple spatially-varying ones, which are automatically learned from the image content itself. Supported by extensive experiments (Sec.~\ref{sec:s4}), MKF outperforms state-of-the-art filters, including BF, TV, and CF, on filtering MR complex images that are corrupted by nonstationary Gaussian noise.

\section{Related Works}
\label{sec:S2}
Image filtering is an ill-posed problem, which needs constraint (according to the prior knowledge, e.g., edge intensity gradient, noise level, or intensity distribution) to improve performance. Early on, for avoiding blurring edges, the mean filter is improved through measuring local statistic; only pixels in the convolution window that has the smallest intensity variance are used to generate the mean~\citep{nagao1979edge}.  %

Afterward, to avoid smoothing across edges that are boundaries, a spatially-invariant weight matrix is employed to detect and extract edges~\citep{burt1981fast}. However, it still blurs edges. Next, the anisotropic diffusion filter employs a Gaussian kernel to precisely detect edges, but it is low in time  efficiency~\citep{perona1990scale}. To overcome such limitations, BF generates spatially-varying weight matrixes at each pixel using the spatially-invariant Gaussian kernel~\citep{tomasi1998bilateral}. The filtering process of BF can be formulated in a generalized intensity space~\citep{barash2002fundamental}; from this geometric perspective, the detection of edges can be interpreted as measuring the distance between pixels.  

Compared with the prior knowledge of noise level or edge gradient, the piece-wise constant intensity distribution can be naturally considered as another sign of image redundancy. Image denoising can be achieved by suppressing the intensity variation. Total variation (TV) targets this goal by minimizing a global cost~\citep{rudin1992nonlinear}, which is weighted by a regularization parameter that trades between the fidelity and smoothness terms. TV does not need to estimate the noise level and thus is commonly used to cope with spatially-varying noise~\citep{eichner2015real, pizzolato2016noise}. And next, Curvature Filter (CF) extends TV allowing more geometric configurations in local~\citep{gong2017curvature}. Although the regularization parameter exerts a function akin to controlling the SNR of the filtered image, how to initialize it is not intuitive. 


\section{Multi-Kernel Filter (MKF)}
\label{sec:S3}
To improve the adaptivity of BF, we propose MKF that adapts filtering kernels according to specific image characteristics. First, we propose a hierarchically clustering algorithm for image context construction (Sec.~\ref{subsec:s31}); and next, we design two terms for automatically adapting filtering kernels (Sec.~\ref{subsec:s32}). 

\subsection{Hierarchically Clustering for Image Context Construction}
\label{subsec:s31}
We design a hierarchically clustering framework taking the inspiration from Gestalt grouping rules and the top-down principle. A whole group of pixels of an image is iteratively split into smaller groups, and each clustering iteration comprises two stages that simulate similarity and proximity Gestalt grouping rules respectively. At last, an image is represented in multi-scale by a hierarchy of large to small coherent clusters.

\subsubsection{Two-Stage Clustering for Multi-Scale Image Representation}
\label{subsec:s311}
The two-stage clustering comprises 
\begin{inparaenum}[(i)]
	\item intensity-based \textit{similarity clustering}, which simulates the Gestalt similarity rule, and 
	\item and the connectedness-based \textit{proximity clustering}, which simulates the Gestalt proximity rule.  
\end{inparaenum}

At the first stage, similarity clustering employs expectation maximization (EM) clustering algorithm which aggregates the input pixels $\left\{I(\vec{x}_1)\ldots I(\vec{x}_m)\ldots I(\vec{x}_n)\right\}$ into two coherent groups with labels $\left\{\ell_1,\ell_2\right\}$. Their intensity distributions are modeled by two Gaussian distributions, which parameters are learned via maximum log-likelihood,
\begin{equation}
\mathop{\arg\max}_{\theta_{t,\,\ell}^*}\sum_{\ell\in\left\{ \ell_1,\ell_2\right\}}p\left(\ell \mid I(\vec{x}_m),\theta_{t,\,\ell}\right)\log p\left(I(\vec{x}_m), \ell \mid \theta_{t,\,\ell}^*\right),
\label{eq:e2}
\end{equation}
\noindent where $\theta_{t,\,\ell}=\left\{\mu_{t,\,\ell} ,\delta_{t,\,\ell} \right\}$ is a parameter set consisting of mean $\mu_{t,\,\ell}$ and standard deviation $\delta_{t,\,\ell}$ of the $\ell$-th cluster at $t$-th layer.

\begin{figure}
	\centering
	\subfigure{  
		\label{fig:sub:f1a}
		\begin{minipage}[t]{0.23\linewidth}  
			\centering  
			\includegraphics[width=1.55in]{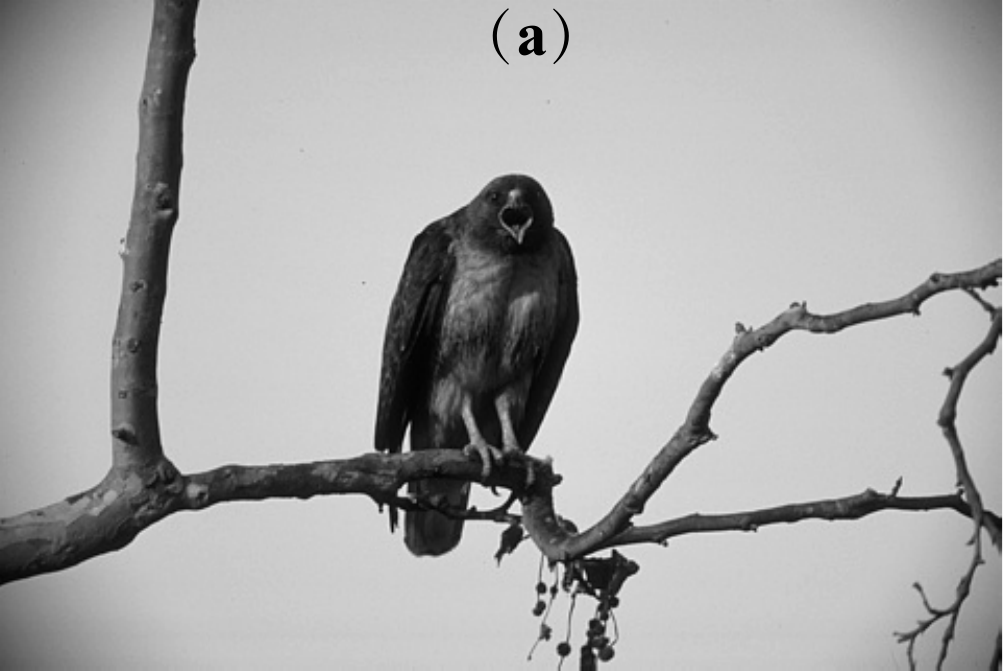}   
		\end{minipage}%
	}%
	\subfigure{  
		\label{fig:sub:f1b}
		\begin{minipage}[t]{0.23\linewidth}  
			\centering  
			\includegraphics[width=1.55in]{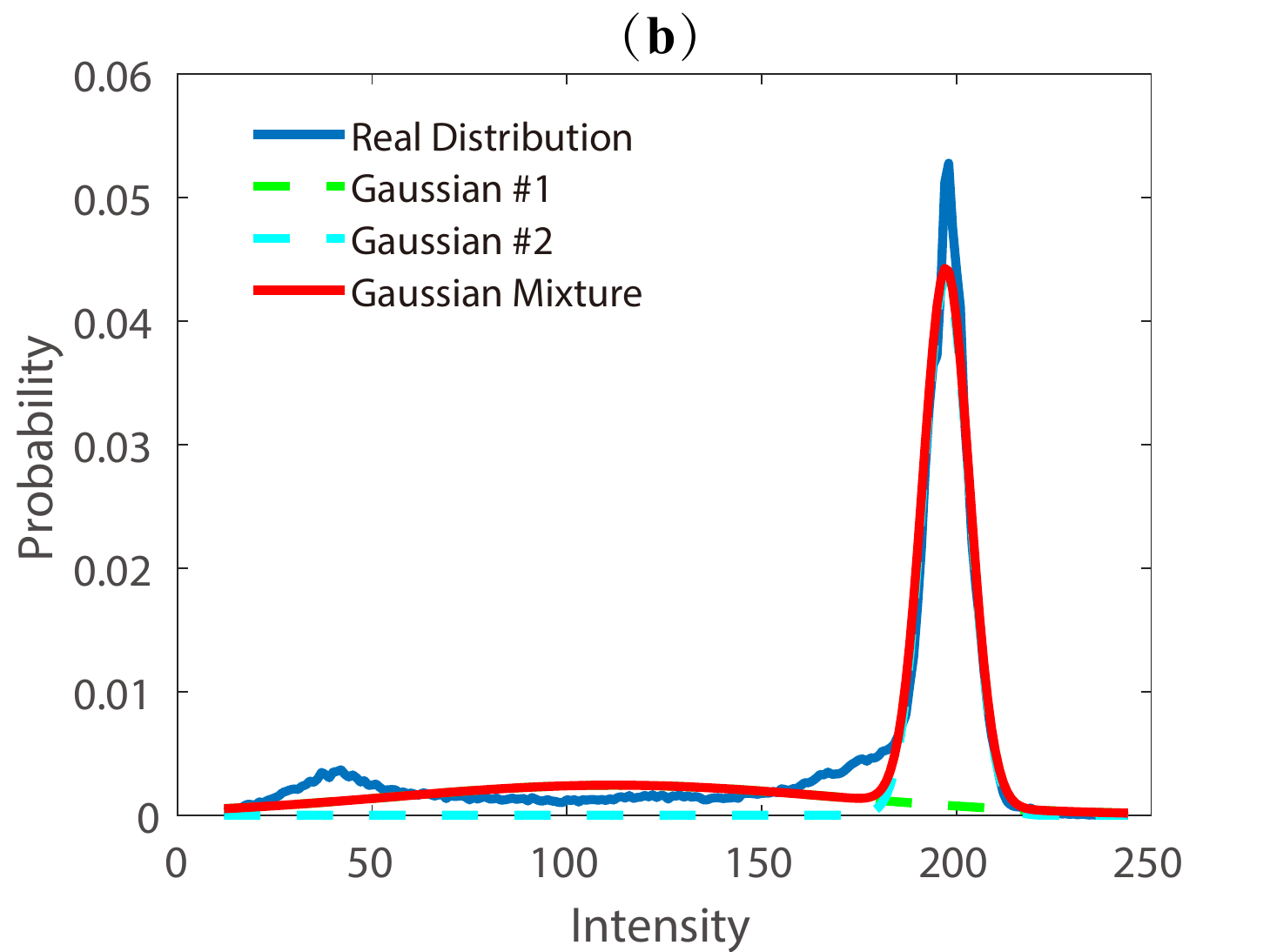}  
		\end{minipage}%
	}%
	\subfigure{  
		\label{fig:sub:f1c}
		\begin{minipage}[t]{0.23\linewidth}  
			\centering  
			\includegraphics[width=1.55in]{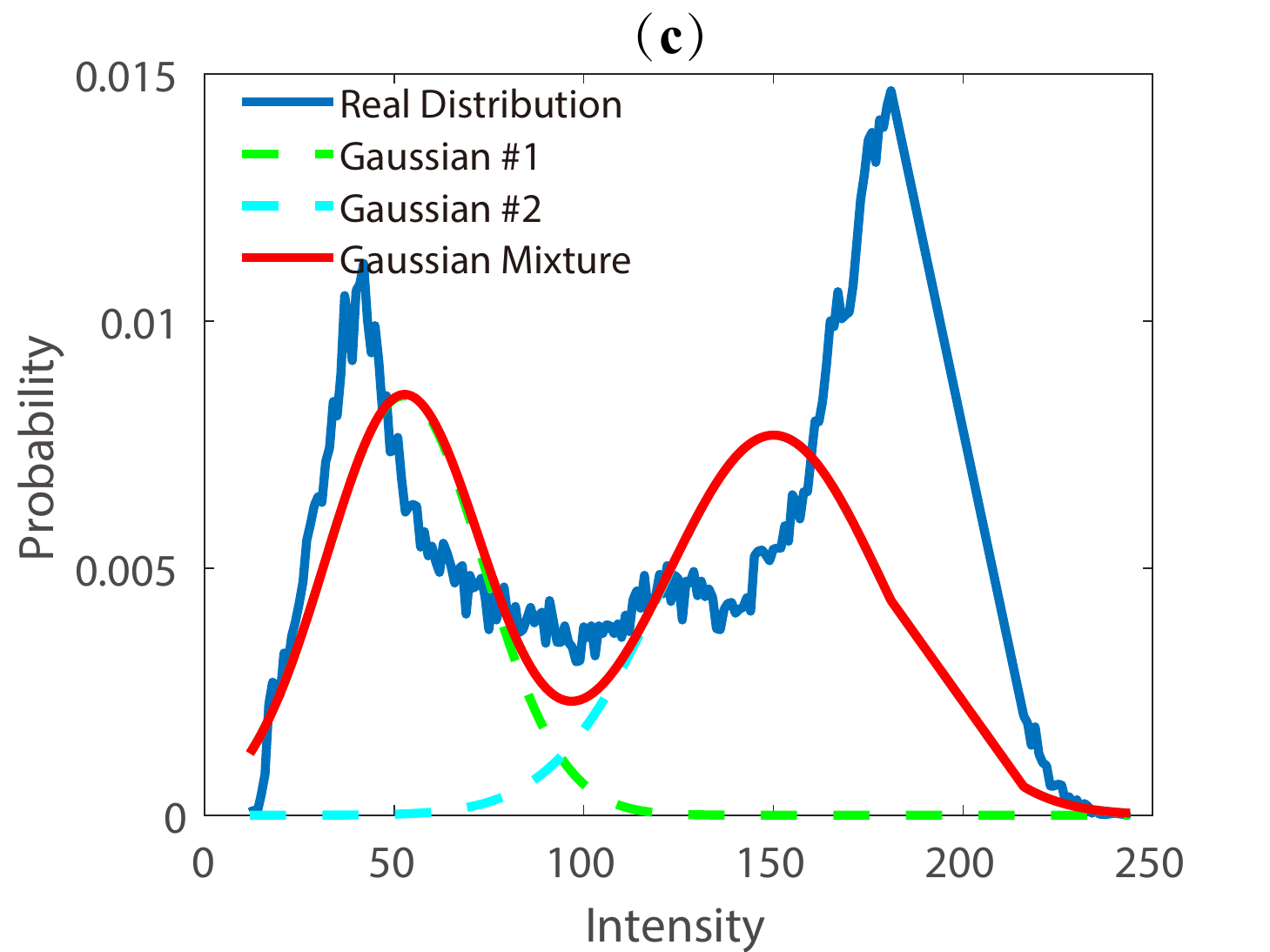}  
		\end{minipage}%
	}%
	\subfigure{ 
		\label{fig:sub:f1d} 
		\begin{minipage}[t]{0.23\linewidth}  
			\centering  
			\includegraphics[width=1.55in]{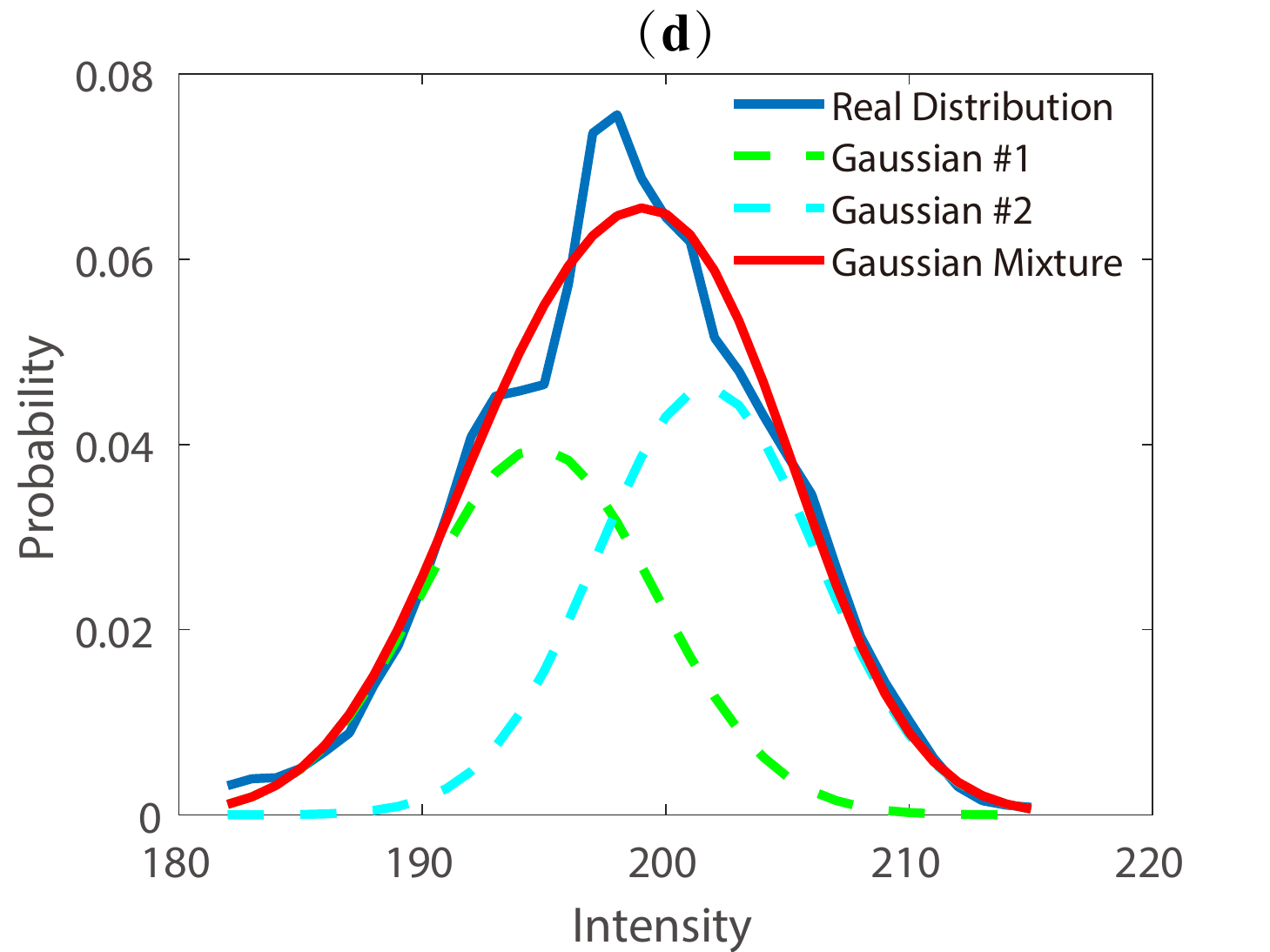}  
		\end{minipage}%
	}
	\caption{Gaussian distributions learned from two rounds of similarity clustering. (a) An eagle image selected from BSD$300$. (b) After the first round, all pixels are aggregated into two groups which intensity distributions are modeled by $\#1$ and $\#2$ Gaussian distributions respectively (denoted by dash-curves). Solid blue curves in (c) and (d) show exact intensity distributions of the two groups of pixels respectively. Each solid red curve denotes the mixture of two Gaussian distributions. }
	\label{fig:F1}
\end{figure}

\begin{figure} 
	\subfigure[Label map of the first round of similarity clustering. Yellow image regions are isolated by the blue tree branches.]{
		\label{fig:sub:f2a}
		\includegraphics[width=0.3\textwidth]{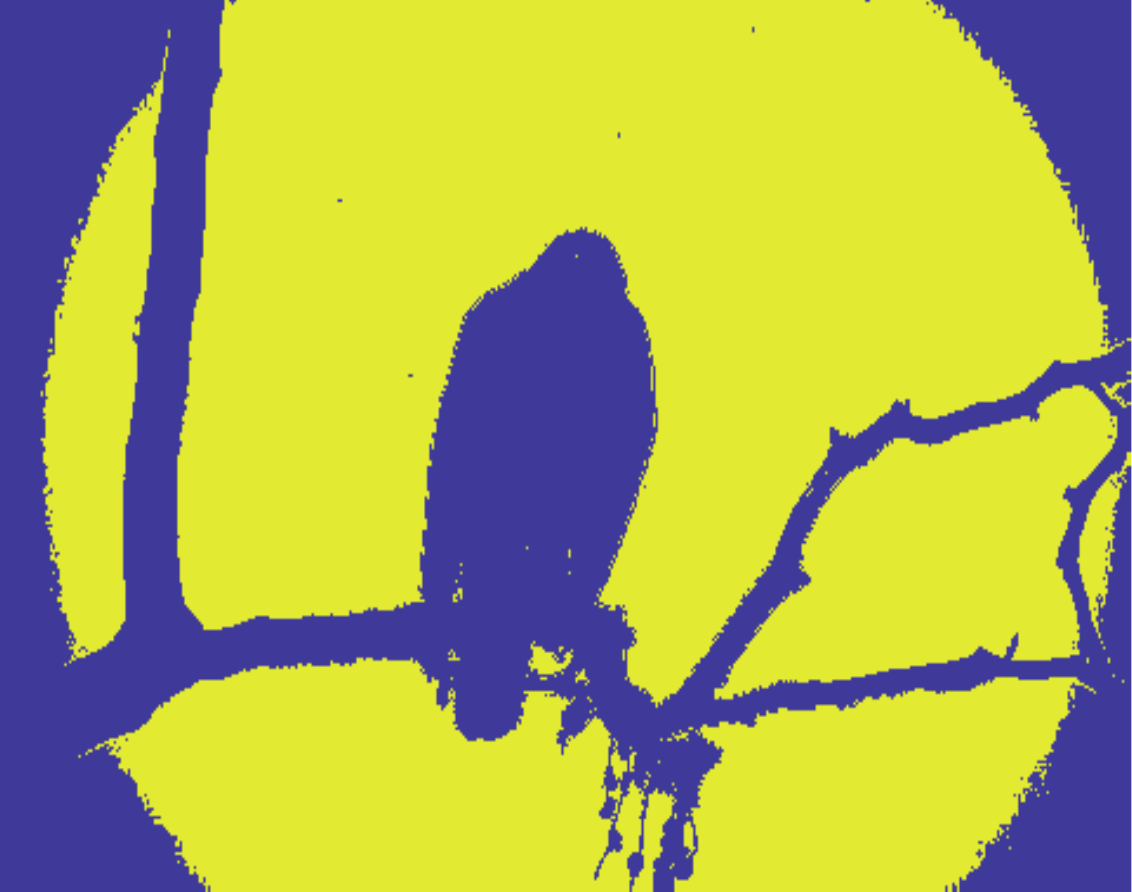}
	}	
	\subfigure[Label map of the second round of similarity clustering (which input is the outcome of the $1^{st}$ similairty clustering rather than proximity clustering )]{
		\label{fig:sub:f2b}
		\includegraphics[width=0.3\textwidth]{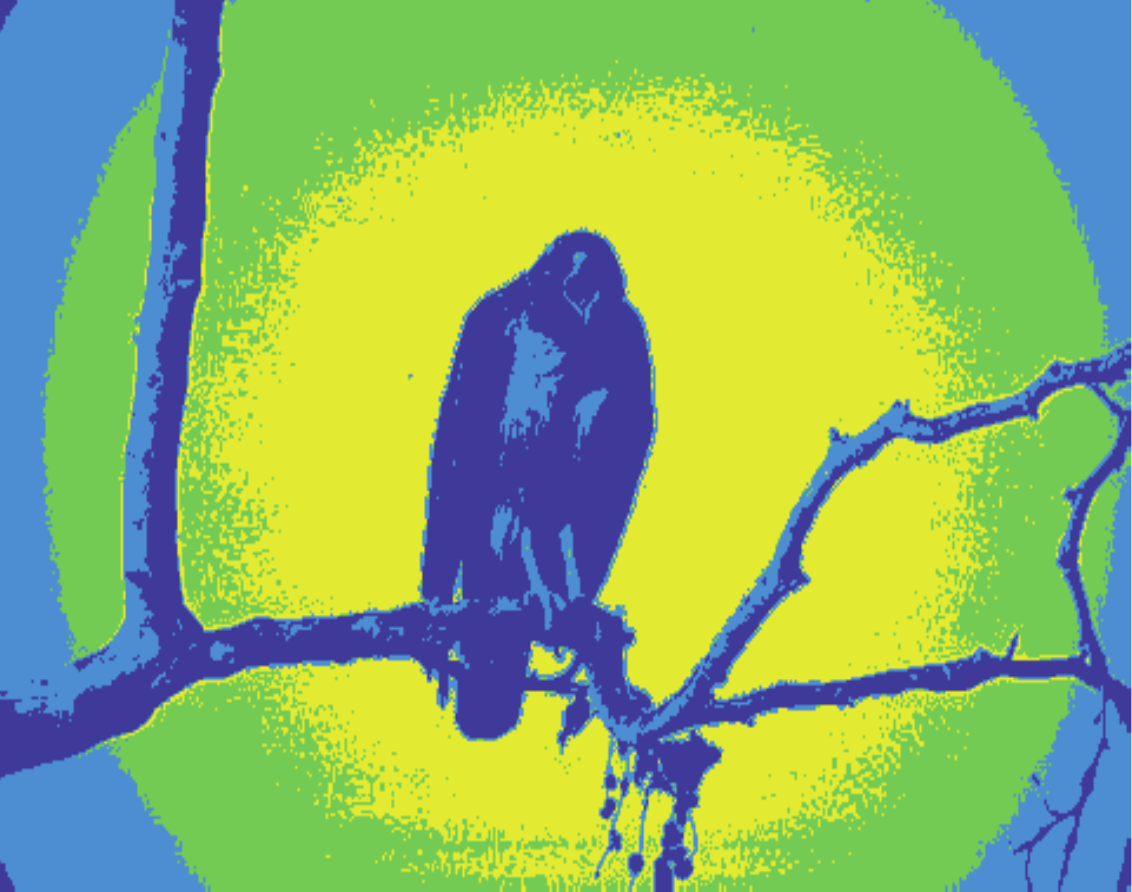}		
	}
    \subfigure[Label map of the first round of proximity clustering. The yellow unconnected image regions are specified unique labels. ]{
    	\label{fig:sub:f2c}
    	\includegraphics[width=0.3\textwidth]{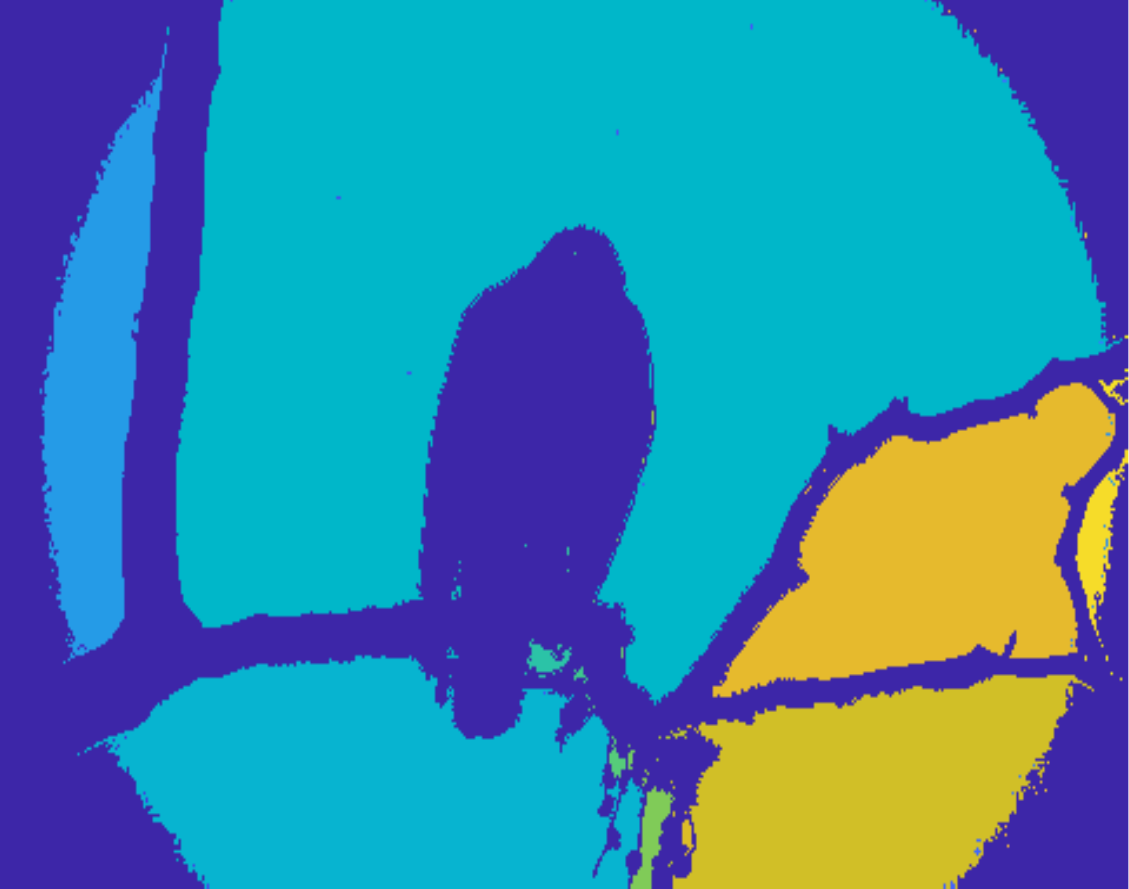}		
    }
	\centering
	\caption{Label maps of intermediate clustering stages. }
	\label{fig:F2}      
\end{figure}

Figure~\ref{fig:F1} shows the Gaussian distributions adopted from two rounds of similarity clustering, without proximity clustering. After the first round of clustering, input image is partitioned into two groups modeled by two Gaussian distributions respectively, shown in Figure~\ref{fig:sub:f1b}. Based on the coarse outcomes, the second round of clustering generates Gaussian distributions in a finer scale, shown in Figure~\ref{fig:sub:f1c} and \ref{fig:sub:f1d} respectively. 
Corresponding label maps are shown in Figure~\ref{fig:F2}.

At the second stage, proximity clustering further isolates unconnected clusters of the outcome of similarity clustering. We can see in Figure \ref{fig:sub:f2a}, the yellow regions are isolated by blue branches; hereby, proximity clustering isolates the unconnected image regions and attaches unique labels to them, shown in Figure \ref{fig:sub:f2c}. As a result, the total number of clusters increase to $n_t$ at $t$-th level, accordingly, the image $I(\vec{x})$ can be represented by Gaussian mixture model (GMM) 
\begin{equation}
\centering
G_t = \sum_{k=1}^{n_t}\mathcal{N}(\mu_{t,\,k},\delta_{t,\,k}),
\label{eq:e3}
\end{equation}
\noindent As $t$ increases, the total number of clusters, $n_t$, increases and cluster inner intensity variation, $\delta_{t,\,k}$, decreases accordingly. Thus, $G_t$ represents the image in fine-scale.

It is noteworthy that the hierarchically clustering framework has the advantage that automatically adapts the total number of clusters to an image. Under images with various characteristics, the fixed number may not suit them appropriately, while the heuristic strategy outperforms conventional flat clustering methods, which initialize a fixed number manually.

\subsubsection{Image Context Represented by Cluster Tree}
\label{subsec:s312}

\begin{figure}
	\centering
	\includegraphics[width=0.83\textwidth]{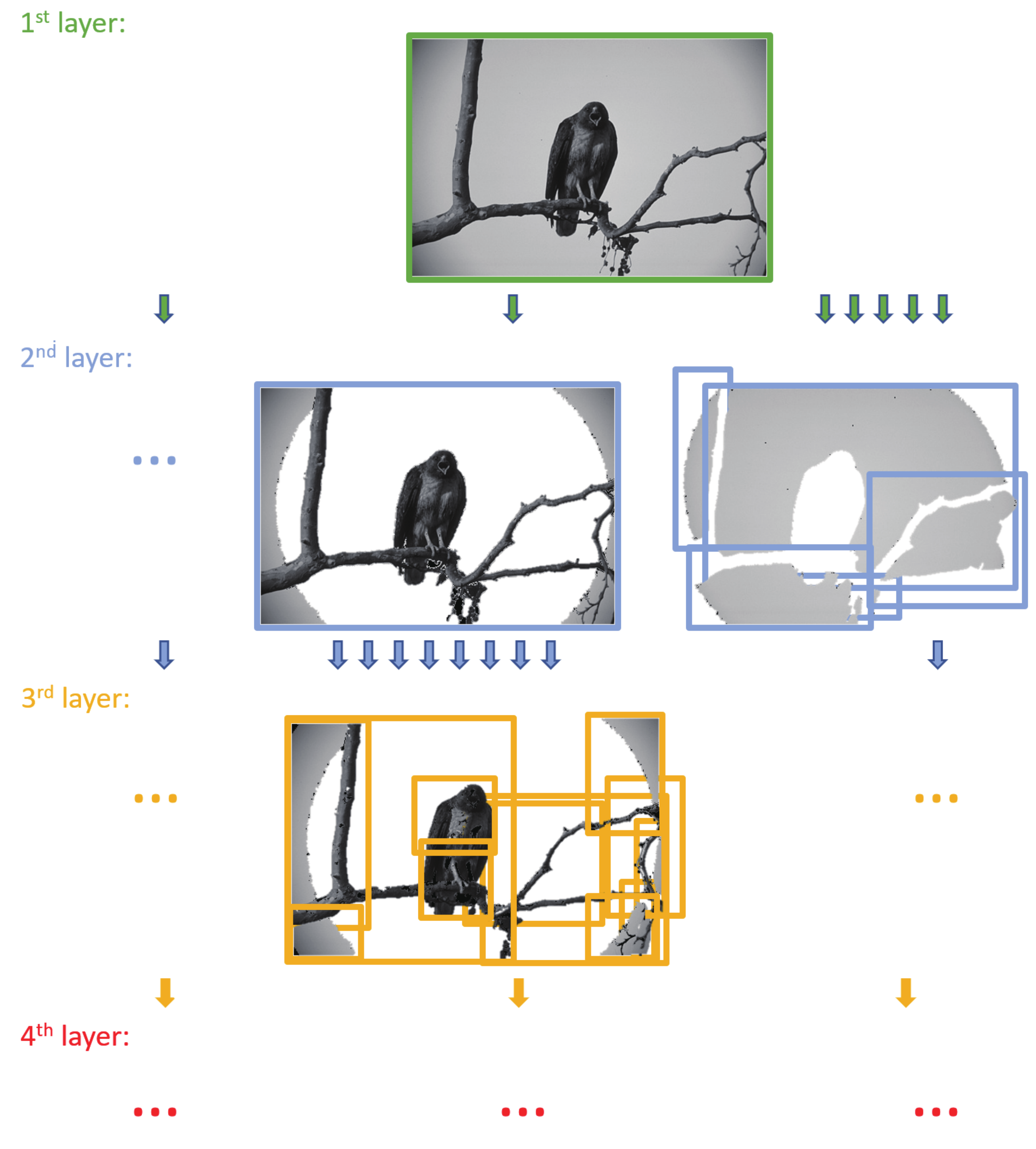}	
	\caption{Partial branches of a cluster tree. Each bounding box denotes a cluster. We can see a cluster in $2^{nd}$ layer contains high-contrast contents, such as the whole eagle; while a cluster in $3^{rd}$ layer only contains one leg which is lower-contrast. We specify a cluster in an upper layer as image context of corresponding leaf clusters.}
	\label{fig:F3}      
\end{figure}

In a human visual system, coarse-scale signals provide contextual cues guiding to form finer-scale vision~\citep{bar2004visual, peterson2013gestalt, lu2018revealing}. Inspired by this, we construct the cluster tree, in which a coarse-level cluster is specified as the image context of its corresponding child clusters in the leaf level. We retrieve the labels of a pixel across levels so that establish the affiliation of clusters across scales. Figure~\ref{fig:F3} vividly illustrates an adopted cluster tree. From the root to leaf, we can find the top to bottom layers of clusters, represent the image from coarse to fine.

To construct the cluster tree, MKF should be initialized by several parameters, which will be introduced in detail below. As for filtering, the deviations, $\delta_{t,\,k}$, in the leaf layer are exploited to automatically generate spatially-varying kernels, while the upper ones bring contextual constraint to fine-tune adopted kernels.

\subsection{Multi-Kernel Filter (MKF) for Image Denoising}
\label{subsec:s32}
BF works effectively when the noise is stationary. MKF extends BF to deal with non-stationary noise (i.e., integrally-varying noise and spatially-varying noise). MKF adapts filtering kernels exploiting image content itself to deal with such noise variability. In this section, we will first reinterpret BF in a generalized intensity space \cite{barash2002fundamental}, and next, we will introduce the extended filtering kernel.  

\subsubsection{Reinterpretation of BF}
\label{subsec:s321}

BF is a weighted-mean filter, given by, 
\begin{equation}
\centering
O\left(I(\vec{x}))\right) = \frac{\sum_{\vec{\xi}}\hat{w}(\vec x, \vec{\xi}) \left(I(\vec{\xi})\right)} {\sum_{\vec{\xi}}\hat{w}(\vec x, \vec{\xi})},
\label{eq:e4}
\end{equation}
\noindent where $\vec{x}$ and $\vec{\xi}$ are the coordinates of center and neighbor pixels respectively, and the filtering kernel
\begin{equation}
\centering
w(\vec{x}, \vec{\xi})= \exp{\left\{ \frac{-(\vec{x}-\vec{\xi})^2}{2h_{\vec{x}}^2 } \right\}} \exp{\left\{\frac{-(I(\vec{x}) - I(\vec{\xi}))^2}{2{h_I}^2} \right\}},
\label{eq:e5}
\end{equation}
\noindent where $\text{h}_{\vec{x}}$ and $\text{h}_I$ are the manually specified parameters in spatial kernel and range kernel respectively. Typically, $h_{\vec{I}}$ involves the prior knowledge of the noise level, and the filtering process can be written in the form
\begin{equation}
\centering
O\left(I(\vec{x})\right)= I(\vec{x}) - R\left(\hat{I}(\vec{x}),\hat{I}(\vec{\xi})\right),
\label{eq:e6}
\end{equation}
\noindent where,
\begin{equation}
\centering
R\left(\hat{I}(\vec{x}),\hat{I}(\vec{\xi})\right) = \frac{\sum_{\vec{\xi}}\hat{w}(\vec x, \vec{\xi})\left(I(\vec{x}) - I(\vec{\xi})\right)} {\sum_{\vec{\xi}}\hat{w}(\vec{x}, \vec{\xi})},
\label{eq:e7}
\end{equation}
\noindent where $I(\vec{\xi})-I(\vec{x})$ is a measure of intensity difference in neighborhood,  $R\left(\hat{I}(\vec{x}),\hat{I}(\vec{\xi})\right)$ estimates and eliminates the local intensity variation caused by noise, which can be analyzed in the generalized intensity space, where the intensity is denoted by,
\begin{equation}
\centering
\hat{I}(\vec{x}) = \left\{\frac{I(\vec{x})}{h_I}, \frac{\vec{x}}{h_{\vec{x}}} \right\}.
\label{eq:e8}
\end{equation}
\noindent Thus, the filtering kernel can be written in the form 
\begin{equation}
\centering
\hat{w}(\vec{x},\vec{\xi})\,=\,\exp{\left\{-\frac{1}{2}\left\|\hat{I}(\vec{x})-\hat{I}(\vec{\xi})\right\|^2\right\}},
\label{eq:e9}
\end{equation}
\noindent where $\|I(\vec{\xi})-I(\vec{x})\|$ is the normalized Euclidean distance in the generalized intensity space. Considering a grey image as the $2$D manifold in $3$D Euclidean space, BF can be interpreted to estimate noise-caused intensity variation through weighting local surface smoothness, because $R\left(\hat{I}(\vec{x}),\hat{I}(\vec{\xi})\right)$ is in the position of $n(\vec{x})$ in Eq.\ref{eq:e1}. With the prior knowledge of the noise level, BF effectively enhances surface smoothness for noise removal. However, with the manually specified parameter of $h_I$, BF cannot carter to noise variability inherently. We hence extend the range kernel of BF.

\subsubsection{Adaptive Filtering of MKF}
\label{subsec:s322}

Following Eq.\ref{eq:e6}, MKF is formulated as,
\begin{equation}
\centering
O\left(I(\vec{x})\right)= I(\vec{x}) - R\left(\hat{I}(\vec{x}), \hat{I}(\vec{\xi}) \mid \delta_{t,\,k},\varPsi_{t,\,k}\right),
\label{eq:e10}
\end{equation}
\noindent where $R(\hat{I}(\vec{x}), \hat{I}(\vec{\xi}) \mid \delta_{t,\,k},\varPsi_{t,\,k})$ generates weights depending not only on  $\delta_{t,\,k}$, which encodes local information, but also on $\varPsi_{t,\,k}$, which takes into consideration contextual information. The filtering kernel employed in $R(\cdot)$ is given by, 
\begin{equation}
\centering
w(\vec{x}, \vec{\xi}\mid \delta_{t,\,k}, \varPsi_{t,\,k})= \exp{\left\{ \frac{-(\vec x-\vec{\xi})^2}{2h_{\vec{x}}^2 } + \frac{-{(I(\vec{x}) - I(\vec{\xi}))}^2\varPsi_{t,\,k}}{2\delta_{t,\,k}^2} \right\}},
\label{eq:e11}
\end{equation}
\noindent where, 
\begin{equation}
\centering
\varPsi_{t,\,k} = {\left( \frac{\delta_{{t-1},\,{k^*}}}{ \delta_{{t-2},\,{k^{**}}}} \right)}^2,
\label{eq:e12}
\end{equation}
\noindent where $k^*$ and $k^{**}$ denote corresponding labels of $k$ in $(t-1)$-th and $(t-2)$-th layers respectively. 

The adaptivity of MKF can be interpreted by the integral function of $\delta_{t,\,k}$ and $\varPsi_{t,\,k}$. Both make the same local intensity gradient $I(\vec{\xi})-I(\vec{x})$ be able to generate different weights. On the other hand, $\varPsi_{t,\,k}$ can be also considered exerting a function on $\delta_{t,\,k}$. That is to say, the same local variation level, $\delta_{t,\,k}$, also means differently, and the specific meaning is constrained by the image context, i.e., $\varPsi_{t,\,k}$. Typically, $\delta_{t-2,\,k^{**}}>delta_{t-1,\,k^{*}}$, therefore, $\varPsi_{t,\,k}<1$ which enlarges $\delta_{t,\,k}$. 

The function of both terms in the filtering kernel of MKF simulates the selective attention mechanisms of human. More specifically, salient objects in circumstance drive human to neglect non-salient objects in local, involuntarily. In MKF, $\varPsi_{t,\,k}$ and $\delta_{t,\,k}$ drive filtering kernels to neglect detailed contents within $k$-th cluster at the leaf level, adaptively. 

Denote a leaf cluster, C$_{t,\,k}$, its context is provided by the clusters in upper levels, i.e., C$_{{t-1},\,{k^*}}$ and C$_{{t-2},\,{k^{**}}}$, which surround C$_{t,\,k}$. A $\varPsi_{t,\,k}$ close to $0$ indicates a high-level salience in surrounding image region, which significantly enlarges $\delta_{t,\,k}$ allowing that MKF smooths across the boundaries of C$_{t,\,k}$. Therefore, lower salient image contents represented by C$_{t,\,k}$ are smoothed out. Otherwise, when $\varPsi_{t,\,k}$ is close to $1$, it denotes low-level salience of surrounding image regions. The $\varPsi_{t,\,k}$ slightly enlarges $\delta_{t,\,k}$ for intraregionally smoothing allowing that the filtering kernel preserves the boundaries of C$_{t,\,k}$. 

Incorporating $\delta_{t,\,k}$ and $\varPsi_{t,\,k}$ into the range kernel, MKF adaptively determines what contents should be smoothed out. Ultimately, the proposed strategy enables MKF to cater  to noise variabilities.

\section{Experiments}
\label{sec:s4} 
The proposed MKF was evaluated on two public datasets, BSD$300$~\citep{MartinFTM01} and BrainWebs~\citep{kwan1999mri}, compared with state-of-the-art filters, including BF~\citep{tomasi1998bilateral},  TV~\citep{rudin1992nonlinear}, CF~\citep{gong2017curvature}. Both datasets were added integrally-varying noise and spatially-varying noise, respectively, according to the noise distribution in practice. 

\subsection{Parameter Setting}
\label{subsec:41}
The initialization of MKF involves $8$ parameters, where the first $3$ ones were specified different values in the following experiments. 

\begin{enumerate}	
	\item Maximal clustering iteration limit, $D_p$, determines the depth of cluster tree. 
	
	\item Maximal cluster size, $M_{x_c}$, determines the maximal number of pixels within a cluster. 

	\item Conduction coefficient, $\text{h}_x$, determines the attenuation rate of weights over spatial. 
	
	\item Minimal cluster size, $M_{n_c}$, determines the lower bound of the cluster size. The clusters with a size smaller or equal $9$ pixels were too small to have meaningful local statistics. They must inherit the local statistic $\theta$ and image context from its eligible parents. 
	
	\item Neighborhood range, $N_e$. We set $N_e$ to $8$ so that proximity clustering examined the connectedness over $8$ neighbor pixels.
	
	\item Intensity distribution precision, $P$, determines the bin width of the intensity histogram that is used to generate Gaussian distributions. We set $P$ to $1$ so that the noise disturbance which value smaller than $1$ was truncated. 
	
	\item In EM clustering algorithm, initial parameter set of Gaussian distributions,  $\theta_\ell^{ini}=\left\{\mu_\ell^{ini}, \delta_\ell^{ini}\right\}$, is the starting condition of each Gaussian distribution to fit the input data. Denote the maximal intensity value as $I_{mx}$, and denote $v$ as the $1$D vector $v=[1,2]$, the parameters were set to, $\mu_\ell^{ini}=v*I_{mx}/(2+1)$, and $\delta_\ell^{ini}=[I_{mx},I_{mx}]$. 
	
	\item Convergence criterion of EM iterations, $T$, was set to $0.0001$.

\end{enumerate}

\subsection{Datasets}
\label{subsec:42}
We fixed well-chosen parameters of filters for observing filtering adaptivity. To show the improvement, MKF was compared with BF on BSD$300$ employing the mean absolute error (MAE) and structural similarity (SSIM) indices. Filters, such as TV and CF, that can cope with spatially-varying noise were only evaluated on BrainWeb. 
 
BSD$300$: Considering the nonlinearity of the chromaticity models, we convert the $100$ test color images to gray-scale. And next, all images are separately added $100$ levels of Gaussian noise, which level is from $10$ to $1000$ with an interval of $10$. The noise level can be normalized by dividing $65025$ ($255\times255$) according to the range of UINT$8$. 

BrainWeb: To generate synthetic MRI data, we employ the noise-free T$1$ volume~\citep{kwan1999mri}. Its intensity range is largely from $-3000$ to $3000$, and we add the spatially-varying Gaussian noise, and the noise level follows $2$D Gaussian distribution. The maximal deviation achieves as large as $500$. Contrasting with the case that image SNR changes with integrally varying noise, the noise levels in synthetic MR data is spatially-varying~\citep{aja2015spatially, chen2018denoising}.

\subsection{Adaptivity under Integrally Varying Noise}
\label{subsec:43}
We first quantitatively evaluated how parameters, including cluster size, $M_{x_c}$, and tree depth, $D_p$, of the clustering impact the smoothing performance of MKF in respect to MAE and SSIM. A small MAE value accompanying a large SSIM value demonstrates high-quality denoising performance.

We chose the $\#33039$ image that was separately added $10$ and $1000$ two levels of noise. To initialize MKF, the tree depth $t$ was chosen from $2$ to $7$, and the cluster size was from $10$ to $200$ with an interval of $10$. Typically, $\varPsi_{t,\,k} = {({\delta_{{t},\,k}} / {\delta_{{t-1},\,{k^{*}}}})}^2$ when $t=2$. 

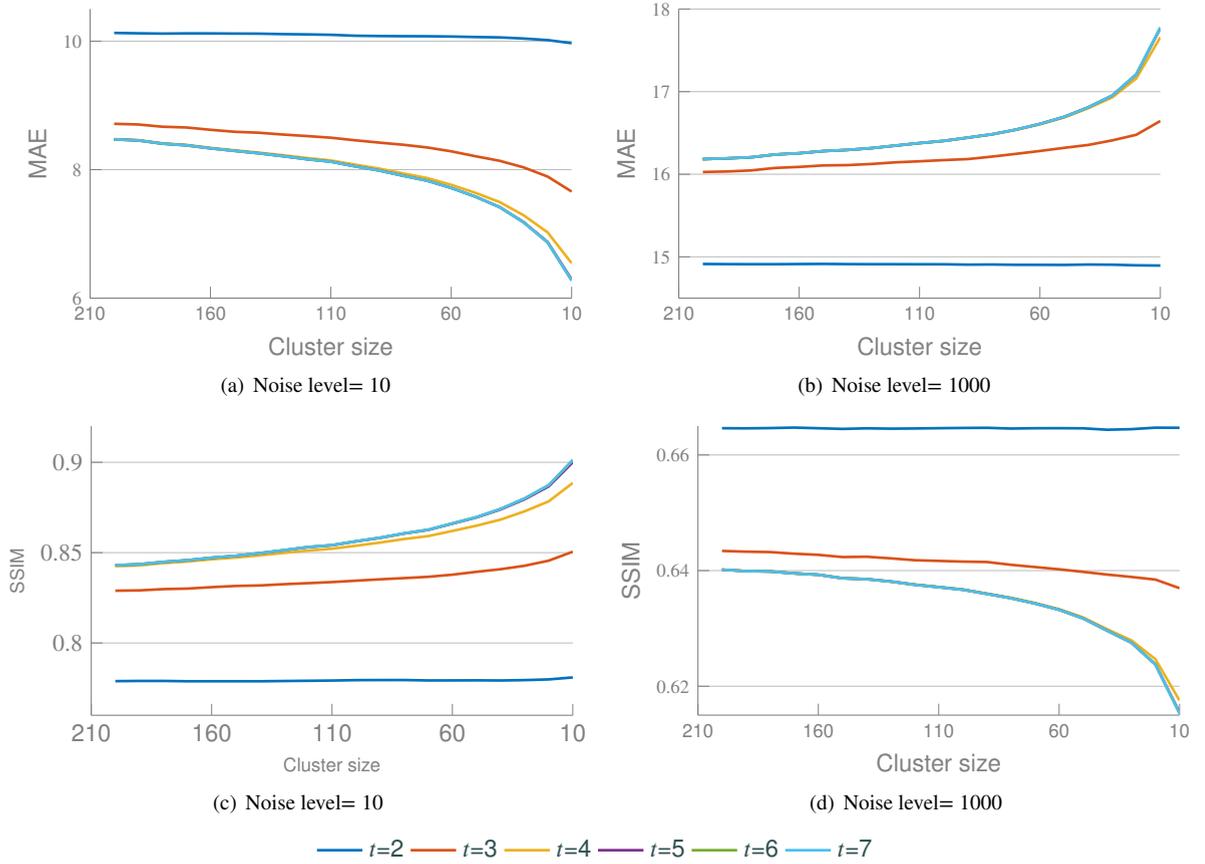
\begin{figure}  
   	\centering	
	\subfigure[Noise level$=10$]{
		\label{fig:sub:f4a}
%
%
\definecolor{mycolor1}{rgb}{0.00000,0.44700,0.74100}%
\definecolor{mycolor2}{rgb}{0.85000,0.32500,0.09800}%
\definecolor{mycolor3}{rgb}{0.92900,0.69400,0.12500}%
\definecolor{mycolor4}{rgb}{0.49400,0.18400,0.55600}%
\definecolor{mycolor5}{rgb}{0.46600,0.67400,0.18800}%
\definecolor{mycolor6}{rgb}{0.30100,0.74500,0.93300}%
\begin{tikzpicture}

\begin{axis}[%
width=2.5in,
height=1.5in,
at={(0.758in,0.481in)},
scale only axis,
xmin=0,
xmax=20,
xtick={0,5,10,15,20},
xticklabels={{210},{160},{110},{60},{10}},
xlabel style={font=\color{white!8!black}},
xlabel={Cluster size},
ymin=6,
ymax=10.5,
ylabel style={font=\color{white!8!black}},
ylabel={MAE},
title style={font=\small},
axis background/.style={fill=white},
axis x line*=bottom,
axis y line*=left,
ymajorgrids,
label style={font=\small},
tick label style={font=\scriptsize},
legend columns=6,
legend style={at={(0.17,0.135)}, font=\small, anchor=south west, legend cell align=left, align=left, fill=none, draw=none}, legend to name={legend:Layer},
clip bounding box=upper bound
]
\addplot [color=mycolor1, line width=1.0pt]
  table[row sep=crcr]{%
1	10.1284072647407\\
2	10.1225500563849\\
3	10.1178743038105\\
4	10.1213078065143\\
5	10.120813286279\\
6	10.1182301522025\\
7	10.1171946922977\\
8	10.11004942198\\
9	10.1056394521857\\
10	10.099216707218\\
11	10.0848688770791\\
12	10.0800244640849\\
13	10.0775713263837\\
14	10.0765545544177\\
15	10.0725796698686\\
16	10.064665712683\\
17	10.058950888544\\
18	10.0415891748364\\
19	10.0171572125143\\
20	9.96860798856143\\
};
\addlegendentry{t=2}

\addplot [color=mycolor2, line width=1.0pt]
  table[row sep=crcr]{%
1	8.71380167483207\\
2	8.70327560023437\\
3	8.66848523878266\\
4	8.65673217412105\\
5	8.62045019843526\\
6	8.58958144450386\\
7	8.57523586044515\\
8	8.54690939566648\\
9	8.52333873305886\\
10	8.49729683621883\\
11	8.45765871456638\\
12	8.42138905131611\\
13	8.38785223697144\\
14	8.34545011569669\\
15	8.28521429938426\\
16	8.20958705761173\\
17	8.13867573596657\\
18	8.03594445330434\\
19	7.8930457794322\\
20	7.6592513242642\\
};
\addlegendentry{t=3}

\addplot [color=mycolor3, line width=1.0pt]
  table[row sep=crcr]{%
1	8.47556801983368\\
2	8.46004553398616\\
3	8.41062270862333\\
4	8.38592255969783\\
5	8.33987945515553\\
6	8.30142271414748\\
7	8.26584638891215\\
8	8.22540567836794\\
9	8.18153073373357\\
10	8.14329658704396\\
11	8.07882136389456\\
12	8.0173238310621\\
13	7.94150464996543\\
14	7.86923907440886\\
15	7.7638278675148\\
16	7.63965050401609\\
17	7.49730799971639\\
18	7.29133137652856\\
19	7.02418991342712\\
20	6.54477239133538\\
};
\addlegendentry{t=4}

\addplot [color=mycolor4, line width=1.0pt]
  table[row sep=crcr]{%
1	8.47127210947524\\
2	8.45457050525199\\
3	8.40462391837088\\
4	8.37860627276438\\
5	8.33186835969821\\
6	8.29255106346953\\
7	8.25483998813863\\
8	8.2105884243264\\
9	8.16325839843376\\
10	8.12464742488936\\
11	8.05252030029463\\
12	7.98753724187605\\
13	7.9059595912316\\
14	7.82856490945759\\
15	7.71544412484382\\
16	7.58091575826805\\
17	7.4194575737002\\
18	7.18486692178972\\
19	6.87417099378516\\
20	6.29134716334001\\
};
\addlegendentry{t=5}

\addplot [color=mycolor5, line width=1.0pt]
  table[row sep=crcr]{%
1	8.47141770742531\\
2	8.45471610320207\\
3	8.40476951632095\\
4	8.37875187071446\\
5	8.33201395764828\\
6	8.2926966614196\\
7	8.25498558608871\\
8	8.21073402227647\\
9	8.16285838697209\\
10	8.12424913812981\\
11	8.05199110299354\\
12	7.98700804457497\\
13	7.90520002721789\\
14	7.82753833071272\\
15	7.71428586330992\\
16	7.57937432345165\\
17	7.4172698907115\\
18	7.18033229262541\\
19	6.86696920301163\\
20	6.27547880385605\\
};
\addlegendentry{t=6}

\addplot [color=mycolor6, line width=1.0pt]
  table[row sep=crcr]{%
1	8.47141770742531\\
2	8.45471610320207\\
3	8.40476951632095\\
4	8.37875187071446\\
5	8.33201395764828\\
6	8.2926966614196\\
7	8.25498558608871\\
8	8.21073402227647\\
9	8.16285838697209\\
10	8.12424913812981\\
11	8.05199110299354\\
12	7.98700804457497\\
13	7.90520002721789\\
14	7.82753833071272\\
15	7.71428586330992\\
16	7.57937924404818\\
17	7.41732885431188\\
18	7.18047816145497\\
19	6.86700082008933\\
20	6.27549297117521\\
};
\addlegendentry{t=7}

\end{axis}
\end{tikzpicture}%
	}
	\subfigure[Noise level$=1000$]{
		\label{fig:sub:f4b}
%
%
\definecolor{mycolor1}{rgb}{0.00000,0.44700,0.74100}%
\definecolor{mycolor2}{rgb}{0.85000,0.32500,0.09800}%
\definecolor{mycolor3}{rgb}{0.92900,0.69400,0.12500}%
\definecolor{mycolor4}{rgb}{0.49400,0.18400,0.55600}%
\definecolor{mycolor5}{rgb}{0.46600,0.67400,0.18800}%
\definecolor{mycolor6}{rgb}{0.30100,0.74500,0.93300}%
\begin{tikzpicture}

\begin{axis}[%
width=2.5in,
height=1.5in,
at={(0.758in,0.481in)},
scale only axis,
xmin=0,
xmax=20,
xtick={0,5,10,15,20},
xticklabels={{210},{160},{110},{60},{10}},
xlabel style={font=\color{white!8!black}},
xlabel={Cluster size},
ymin=14.5,
ymax=18,
ylabel style={font=\color{white!8!black}},
ylabel={MAE},
title style={font=\small},
axis background/.style={fill=white},
axis x line*=bottom,
axis y line*=left,
ymajorgrids,
label style={font=\small},
tick label style={font=\scriptsize},
legend columns=6,
legend style={at={(0.168,0.645)}, font=\small, anchor=south west, legend cell align=left, align=left, fill=none, draw=none}, legend to name={legend:Layer},
clip bounding box=upper bound
]
\addplot [color=mycolor1, line width=1.0pt]
  table[row sep=crcr]{%
1	14.9135106929349\\
2	14.9119292899263\\
3	14.9109076709355\\
4	14.9110544507129\\
5	14.9129192361133\\
6	14.9144899518978\\
7	14.9120084317979\\
8	14.9111108777703\\
9	14.9108627622989\\
10	14.9100811343138\\
11	14.9098609864986\\
12	14.9056477496213\\
13	14.9073327127496\\
14	14.9035724601364\\
15	14.9032817806249\\
16	14.9021849043043\\
17	14.9069171102622\\
18	14.9050185143942\\
19	14.8973295802666\\
20	14.894537492401\\
};
\addlegendentry{t=2}

\addplot [color=mycolor2, line width=1.0pt]
  table[row sep=crcr]{%
1	16.02737365629\\
2	16.0332392547014\\
3	16.0462084958489\\
4	16.0761960015972\\
5	16.0888462638936\\
6	16.1067774459635\\
7	16.1104245683519\\
8	16.1243867868575\\
9	16.1441105153272\\
10	16.1568040853717\\
11	16.171168952446\\
12	16.1833278317232\\
13	16.2131512166974\\
14	16.2466491929014\\
15	16.2817114437838\\
16	16.3193276392536\\
17	16.3543833316621\\
18	16.4098265002869\\
19	16.4777497488503\\
20	16.6447237474721\\
};
\addlegendentry{t=3}

\addplot [color=mycolor3, line width=1.0pt]
  table[row sep=crcr]{%
1	16.1830830711083\\
2	16.1920337661213\\
3	16.2050030072688\\
4	16.238685149101\\
5	16.2542630197068\\
6	16.2793641098144\\
7	16.2935011078995\\
8	16.3152254581126\\
9	16.3455525425693\\
10	16.376068184775\\
11	16.4022795880554\\
12	16.4415987606684\\
13	16.480953821387\\
14	16.5360420765553\\
15	16.6042065543466\\
16	16.6865103012711\\
17	16.7986252363114\\
18	16.9299928150549\\
19	17.1596396770282\\
20	17.6577599709487\\
};
\addlegendentry{t=4}

\addplot [color=mycolor4, line width=1.0pt]
  table[row sep=crcr]{%
1	16.1830830711083\\
2	16.1920337661213\\
3	16.2050030072688\\
4	16.238685149101\\
5	16.2542630197068\\
6	16.2802104194531\\
7	16.2943474175382\\
8	16.3160717677513\\
9	16.3467717709969\\
10	16.3772874132026\\
11	16.4039078952917\\
12	16.4432270679047\\
13	16.4834502971409\\
14	16.5400555359711\\
15	16.6094862783871\\
16	16.6944382626266\\
17	16.8095106625541\\
18	16.9519553785827\\
19	17.207659314536\\
20	17.7681232230388\\
};
\addlegendentry{t=5}

\addplot [color=mycolor5, line width=1.0pt]
  table[row sep=crcr]{%
1	16.1830830711083\\
2	16.1920337661213\\
3	16.2050030072688\\
4	16.238685149101\\
5	16.2542630197068\\
6	16.2802104194531\\
7	16.2943474175382\\
8	16.3160717677513\\
9	16.3467717709969\\
10	16.3772874132026\\
11	16.4039078952917\\
12	16.4432270679047\\
13	16.4834502971409\\
14	16.5400555359711\\
15	16.6094862783871\\
16	16.6944382626266\\
17	16.8095325314168\\
18	16.9519772474454\\
19	17.2085140268887\\
20	17.7752363211685\\
};
\addlegendentry{t=6}

\addplot [color=mycolor6, line width=1.0pt]
  table[row sep=crcr]{%
1	16.1830830711083\\
2	16.1920337661213\\
3	16.2050030072688\\
4	16.238685149101\\
5	16.2542630197068\\
6	16.2802104194531\\
7	16.2943474175382\\
8	16.3160717677513\\
9	16.3467717709969\\
10	16.3772874132026\\
11	16.4039078952917\\
12	16.4432270679047\\
13	16.4834502971409\\
14	16.5400555359711\\
15	16.6094862783871\\
16	16.6944382626266\\
17	16.8095325314168\\
18	16.9519772474454\\
19	17.2085140268887\\
20	17.7753644054263\\
};
\addlegendentry{t=7}

\end{axis}
\end{tikzpicture}%
	} 
	\subfigure[Noise level$=10$]{
		\label{fig:sub:f4c}
%
%
\definecolor{mycolor1}{rgb}{0.00000,0.44700,0.74100}%
\definecolor{mycolor2}{rgb}{0.85000,0.32500,0.09800}%
\definecolor{mycolor3}{rgb}{0.92900,0.69400,0.12500}%
\definecolor{mycolor4}{rgb}{0.49400,0.18400,0.55600}%
\definecolor{mycolor5}{rgb}{0.46600,0.67400,0.18800}%
\definecolor{mycolor6}{rgb}{0.30100,0.74500,0.93300}%
\begin{tikzpicture}

\begin{axis}[%
width=2.5in,
height=1.5in,
at={(0.758in,0.481in)},
scale only axis,
xmin=0,
xmax=20,
xtick={0,5,10,15,20},
xticklabels={{210},{160},{110},{60},{10}},
xlabel style={font=\color{white!8!black}},
xlabel={Cluster size},
ymin=0.76,
ymax=0.92,
ylabel style={font=\color{white!8!black}},
ylabel={SSIM},
title style={font=\small},
axis background/.style={fill=white},
axis x line*=bottom,
axis y line*=left,
ymajorgrids,
label style={font=\scriptsize},
tick label style={font=\small},
legend columns=6,
legend style={at={(0.179,0.671)}, font=\small, anchor=south west, legend cell align=left, align=left, fill=none, draw=none}, legend to name={legend:Layer},
clip bounding box=upper bound
]
\addplot [color=mycolor1, line width=1.0pt]
  table[row sep=crcr]{%
1	0.778888244152038\\
2	0.778997018696173\\
3	0.77899744775544\\
4	0.778812042467943\\
5	0.778823185111145\\
6	0.778823847140529\\
7	0.778812188242946\\
8	0.778955952429968\\
9	0.779077537890729\\
10	0.77920936913058\\
11	0.779444860989557\\
12	0.779480238250901\\
13	0.779501921087387\\
14	0.779267289294587\\
15	0.779284234999385\\
16	0.779323228282458\\
17	0.779244193911929\\
18	0.779458052428303\\
19	0.779845058767643\\
20	0.780936935548983\\
};
\addlegendentry{$t$=2}

\addplot [color=mycolor2, line width=1.0pt]
  table[row sep=crcr]{%
1	0.828927894813367\\
2	0.829111923318194\\
3	0.829813733336244\\
4	0.83010444593094\\
5	0.83090010825552\\
6	0.83156712393645\\
7	0.831837166773111\\
8	0.832528603143362\\
9	0.833103778946452\\
10	0.833718649021163\\
11	0.834480784674854\\
12	0.835242117898308\\
13	0.835916677339012\\
14	0.836627240047104\\
15	0.837806957883948\\
16	0.839391016625795\\
17	0.840833285963892\\
18	0.842749701256994\\
19	0.84555974884379\\
20	0.85052596322736\\
};
\addlegendentry{$t$=3}

\addplot [color=mycolor3, line width=1.0pt]
  table[row sep=crcr]{%
1	0.842487402360314\\
2	0.842935212166368\\
3	0.844205711522292\\
4	0.845091428743422\\
5	0.84632039351464\\
6	0.847263652207831\\
7	0.848491991860501\\
8	0.849844121930699\\
9	0.851150329553953\\
10	0.85216657354202\\
11	0.853833047747887\\
12	0.855583770228693\\
13	0.857536716497295\\
14	0.859166823662904\\
15	0.862004239788541\\
16	0.864923537278751\\
17	0.868286701188064\\
18	0.872879228553714\\
19	0.878380216589836\\
20	0.888587436828879\\
};
\addlegendentry{$t$=4}

\addplot [color=mycolor4, line width=1.0pt]
  table[row sep=crcr]{%
1	0.843009403630856\\
2	0.8435504819714\\
3	0.844849789026749\\
4	0.845785292147968\\
5	0.847168514128136\\
6	0.848209007514301\\
7	0.84969742129306\\
8	0.851315529178231\\
9	0.853031705697697\\
10	0.854073017150309\\
11	0.856222601838631\\
12	0.858177972312141\\
13	0.860581356945471\\
14	0.862580899698116\\
15	0.866039432060336\\
16	0.869465032973493\\
17	0.873894950916964\\
18	0.879516303380682\\
19	0.8866660468352\\
20	0.900008443971483\\
};
\addlegendentry{$t$=5}

\addplot [color=mycolor5, line width=1.0pt]
  table[row sep=crcr]{%
1	0.84304829239282\\
2	0.843589370595278\\
3	0.844888679600123\\
4	0.845824185283653\\
5	0.847207407720219\\
6	0.848247798845879\\
7	0.849735524457101\\
8	0.851353634406944\\
9	0.853098518910688\\
10	0.854161504082695\\
11	0.856322360372578\\
12	0.858277414584766\\
13	0.86069609865574\\
14	0.862715164784065\\
15	0.866196425058554\\
16	0.869670364375806\\
17	0.874209189908549\\
18	0.879993849340609\\
19	0.887340773454572\\
20	0.901177691385851\\
};
\addlegendentry{$t$=6}

\addplot [color=mycolor6, line width=1.0pt]
  table[row sep=crcr]{%
1	0.84304829239282\\
2	0.843589370595278\\
3	0.844888679600123\\
4	0.845824185283653\\
5	0.847207407720219\\
6	0.848247798845879\\
7	0.849735524457101\\
8	0.851353634406944\\
9	0.853098518910688\\
10	0.854161504082695\\
11	0.856322360372578\\
12	0.858277414584766\\
13	0.86069609865574\\
14	0.862715164784065\\
15	0.866196425058554\\
16	0.869674763697866\\
17	0.874211443370503\\
18	0.87999970159276\\
19	0.887349736229791\\
20	0.901246796567479\\
};
\addlegendentry{$t$=7}

\end{axis}
\end{tikzpicture}%
	}
	\subfigure[Noise level$=1000$]{
		\label{fig:sub:f4d}
%
%
\definecolor{mycolor1}{rgb}{0.00000,0.44700,0.74100}%
\definecolor{mycolor2}{rgb}{0.85000,0.32500,0.09800}%
\definecolor{mycolor3}{rgb}{0.92900,0.69400,0.12500}%
\definecolor{mycolor4}{rgb}{0.49400,0.18400,0.55600}%
\definecolor{mycolor5}{rgb}{0.46600,0.67400,0.18800}%
\definecolor{mycolor6}{rgb}{0.30100,0.74500,0.93300}%
\begin{tikzpicture}

\begin{axis}[%
width=2.5in,
height=1.5in,
at={(0.758in,0.481in)},
scale only axis,
xmin=0,
xmax=20,
xtick={0,5,10,15,20},
xticklabels={{210},{160},{110},{60},{10}},
xlabel style={font=\color{white!8!black}},
xlabel={Cluster size},
ymin=0.615,
ymax=0.665,
ylabel style={font=\color{white!8!black}},
ylabel={SSIM},
axis background/.style={fill=white},
axis x line*=bottom,
axis y line*=left,
title style={font=\small},
ymajorgrids,
label style={font=\small},
tick label style={font=\scriptsize},
legend columns=6,
legend style={at={(0.167,0.169)}, font=\small, anchor=south west, legend cell align=left, align=left, fill=none, draw=none}, legend to name={legend:Layer},
clip bounding box=upper bound
]
\addplot [color=mycolor1, line width=1.0pt]
  table[row sep=crcr]{%
1	0.664626174131965\\
2	0.664603941682312\\
3	0.664648787346301\\
4	0.664734009037445\\
5	0.664622183104603\\
6	0.664505104212561\\
7	0.66460434914358\\
8	0.664548710000165\\
9	0.664580503756939\\
10	0.664632254341919\\
11	0.664669812125697\\
12	0.66469649077638\\
13	0.664565415743513\\
14	0.664621546164468\\
15	0.664630857082266\\
16	0.664608429079918\\
17	0.664359589523268\\
18	0.664460395651492\\
19	0.6647211614664\\
20	0.664700121044854\\
};
\addlegendentry{$t$=2}

\addplot [color=mycolor2, line width=1.0pt]
  table[row sep=crcr]{%
1	0.643403305845668\\
2	0.643270218187602\\
3	0.643203827387638\\
4	0.642923715865284\\
5	0.642724424866594\\
6	0.6423419192583\\
7	0.642396214304716\\
8	0.642120383438619\\
9	0.641795353375918\\
10	0.641674946263045\\
11	0.641555036654025\\
12	0.641486323917347\\
13	0.641004862097339\\
14	0.640613435490688\\
15	0.640225049528078\\
16	0.639775873275135\\
17	0.63931092058926\\
18	0.638903445127271\\
19	0.638446062336272\\
20	0.636961105889261\\
};
\addlegendentry{$t$=3}

\addplot [color=mycolor3, line width=1.0pt]
  table[row sep=crcr]{%
1	0.640161980599213\\
2	0.639913854227783\\
3	0.639847531039737\\
4	0.639522221223698\\
5	0.63928315435037\\
6	0.638698298917595\\
7	0.638542898048385\\
8	0.638133551526921\\
9	0.637599368737484\\
10	0.637175948974029\\
11	0.636754323758366\\
12	0.63603507273436\\
13	0.635305558002757\\
14	0.634407642504253\\
15	0.63334262703772\\
16	0.631899309830258\\
17	0.629857011995799\\
18	0.627942476366919\\
19	0.624679255464853\\
20	0.617545490794451\\
};
\addlegendentry{$t$=4}

\addplot [color=mycolor4, line width=1.0pt]
  table[row sep=crcr]{%
1	0.640161980599213\\
2	0.639913854227783\\
3	0.639847531039737\\
4	0.639522221223698\\
5	0.63928315435037\\
6	0.638675891089843\\
7	0.638520490299194\\
8	0.638111143857471\\
9	0.637558589994095\\
10	0.637135171561311\\
11	0.636697844141611\\
12	0.635978591860416\\
13	0.635232329969138\\
14	0.634307961854904\\
15	0.633227244536373\\
16	0.631725692785334\\
17	0.629614077244659\\
18	0.627540932228244\\
19	0.623781243237372\\
20	0.615471793686829\\
};
\addlegendentry{$t$=5}

\addplot [color=mycolor5, line width=1.0pt]
  table[row sep=crcr]{%
1	0.640161980599213\\
2	0.639913854227783\\
3	0.639847531039737\\
4	0.639522221223698\\
5	0.63928315435037\\
6	0.638675891089843\\
7	0.638520490299194\\
8	0.638111143857471\\
9	0.637558589994095\\
10	0.637135171561311\\
11	0.636697844141611\\
12	0.635978591860416\\
13	0.635232329969138\\
14	0.634307961854904\\
15	0.633227244536373\\
16	0.631725692785334\\
17	0.629610313348537\\
18	0.627537167829281\\
19	0.623763487562284\\
20	0.615337053908578\\
};
\addlegendentry{$t$=6}

\addplot [color=mycolor6, line width=1.0pt]
  table[row sep=crcr]{%
1	0.640161980599213\\
2	0.639913854227783\\
3	0.639847531039737\\
4	0.639522221223698\\
5	0.63928315435037\\
6	0.638675891089843\\
7	0.638520490299194\\
8	0.638111143857471\\
9	0.637558589994095\\
10	0.637135171561311\\
11	0.636697844141611\\
12	0.635978591860416\\
13	0.635232329969138\\
14	0.634307961854904\\
15	0.633227244536373\\
16	0.631725692785334\\
17	0.629610313348537\\
18	0.627537167829281\\
19	0.623763487562284\\
20	0.615334234196864\\
};
\addlegendentry{$t$=7}

\end{axis}
\end{tikzpicture}%
	} 
    \definecolor{mycolor1}{rgb}{0.00000,0.44700,0.74100}%
    \definecolor{mycolor2}{rgb}{0.85000,0.32500,0.09800}%
    \definecolor{mycolor3}{rgb}{0.92900,0.69400,0.12500}%
    \definecolor{mycolor4}{rgb}{0.49400,0.18400,0.55600}%
    \definecolor{mycolor5}{rgb}{0.46600,0.67400,0.18800}%
    \definecolor{mycolor6}{rgb}{0.30100,0.74500,0.93300}%
	\ref{legend:Layer}
	\label{fig:F4}  
     \caption{While a large tree depth performs well on the image with small-level noise, a small tree depth is optimal to restore image from severe noise, demonstrating a negative correlation between the tree depth and noise level. } 
\end{figure}

Comparing Figure~\ref{fig:sub:f4a} to Figure~\ref{fig:sub:f4b}, or comparing Figure~\ref{fig:sub:f4c} to Figure~\ref{fig:sub:f4d}, tree depth demonstrated the contrary performance. When $t=7$, MKF achieved the best performance under small level noise while achieving the worst performance under severe noise. In Figure~\ref{fig:F4}, we can also find that the MAE and SSIM curves are flatter when $t=2$, no matter under the noise level of $10$ or $1000$. Typically, MKF performed slightly better when the cluster size became small. 

The contrary effect of tree depth largely owes to the clustering. The depth of cluster tree indirectly affected the cluster size. A small $t$ generated large-size clusters, while a large $t$ made clusters in the leaf layer be small-size. The small cluster represents the image in fine-scale, and hereby, allows MKF to preserve the detailed image contents precisely. On the contrary, severe noise corrupts image contents significantly. The statistic of a small image region is not reliable; while a large image region is preferred owing to more redundant pixels. 

Using the cat image~\citep{tomasi1998bilateral, barash2002fundamental}, which contains both flat and texture regions, we first evaluated the weight matrixes derived by BF and MKF respectively. The $\text{h}_I$ of BF was initialized as $57$, which was the best chosen; $\text{h}_x$ was $3$, and windows radius was $5$. In MKF, we specified the tree depth $2$, cluster size $20$, and other parameters were the same with BF.

\begin{figure} 
	\centering
	\includegraphics[width=1\textwidth]{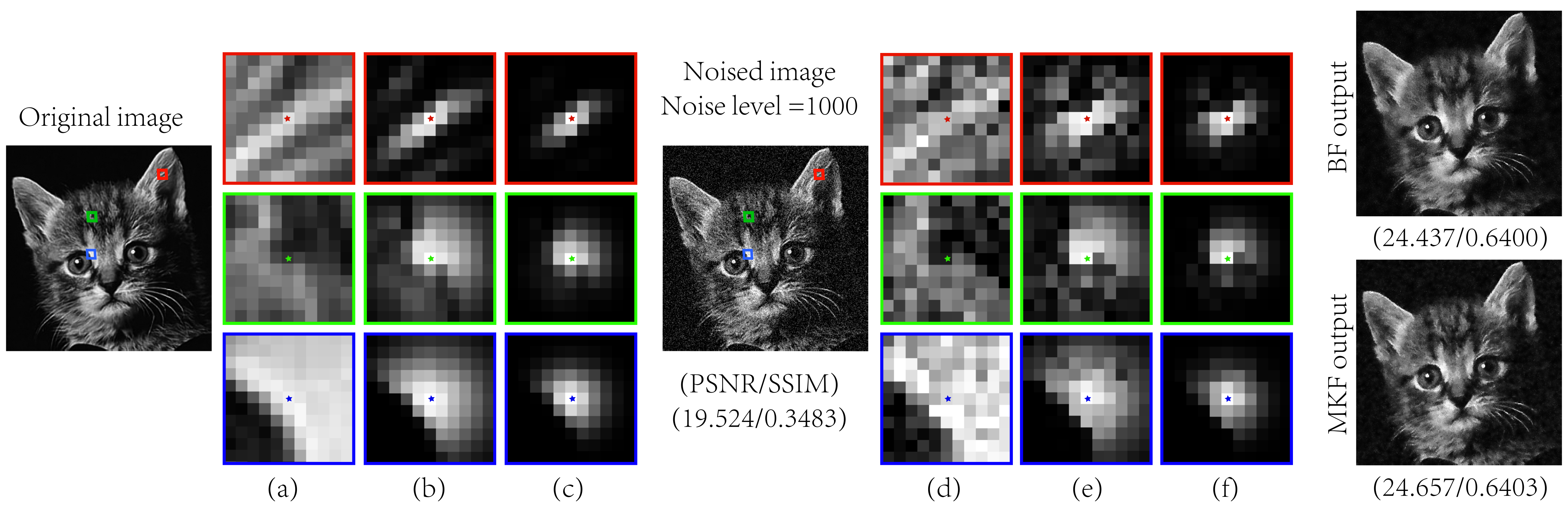}
	\caption{Severe noise ($1000$) significantly corrupts image contents as well as the filtering kernel of BF. Three typical noise-free and noisy image patches, shown in (a) and (d) respectively, are used to demonstrate such degeneration. The filtering kernels of BF are shown in (b) and (e); while (c) and (f) denote MKF. Since a local image region is easily corrupted, BF kernels degenerate more significantly. Using contextual information, MKF demonstrated more robust kernels, allowing to restore more textures and to show better quantitative performance than BF.} 
	\label{fig:F5}      
\end{figure}

The weight matrixes obtained from noise-free and severely noisy images are shown in Figure~\ref{fig:F5}. The weight matrixes of BF were serverly corrupted resulting in the over smoothed cat shown in Figure~\ref{fig:F5}. In the meanwhile, MKF did not only demonstrate more robust filtering kernels but also restore a cat with more detailed textures than BF. 

\begin{figure}
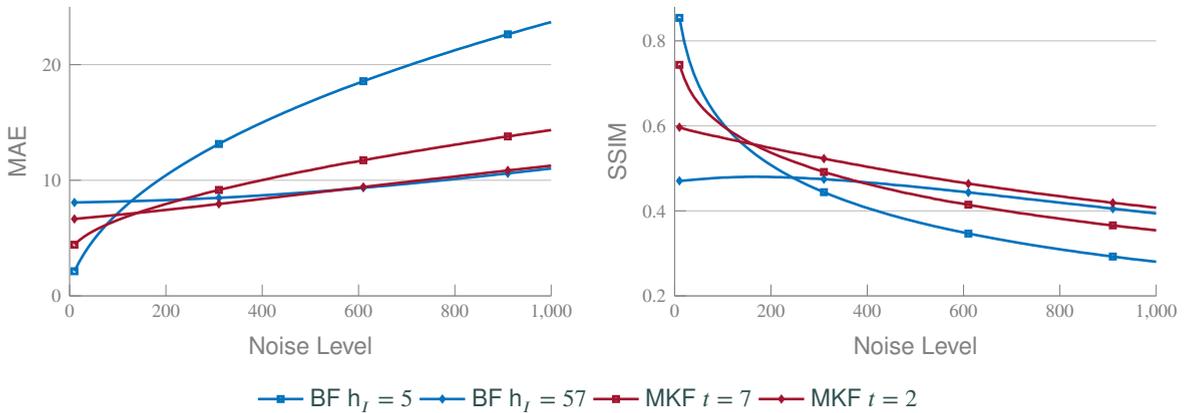
 
	\centering
	\subfigure{
		\label{fig:sub:f6a}
		\subimport{/}{mae_bf_mkf.tex} 
	}
	\subfigure{
		\label{fig:sub:f6b}
		\subimport{/}{ssim_bf_mkf.tex}
	}
    \definecolor{mycolor1}{rgb}{0.85000,0.32500,0.09800}
    \definecolor{mycolor2}{rgb}{0.49400,0.18400,0.55600}
    \definecolor{mycolor3}{rgb}{0.63500,0.07800,0.18400}
    \definecolor{mycolor4}{rgb}{0.30100,0.74500,0.93300}
    \definecolor{mycolor5}{rgb}{0.46600,0.67400,0.18800}
    \definecolor{mycolor6}{rgb}{0.00000,0.44700,0.74100}
    \definecolor{mycolor7}{rgb}{0.92900,0.69400,0.12500}
	\ref{legend:Comp}
	\caption{The curve flatness demonstrates the robustness of filtering. Due to the parameter learning, MKF adapted to noise variabilities better than BF.} 
	\label{fig:F6}       

    \end{figure}

To observe the adaptivity under a series of integrally varying noises over various images, we further evaluated MKF and BF on BSD$300$ under two sets of parameters. For MKF,  
\begin{inparaenum}[(i)]
	\item $D_p=2$, $M_{x_c}=20$; and
	\item $D_p=7$, $M_{x_c}=20$.
\end{inparaenum} 
As for BF,  
\begin{inparaenum}[(i)]
	\item $\text{h}_I = 57$; and
	\item $\text{h}_I = 5$.
\end{inparaenum} 
Other parameters were the same as the above experiment. As shown in Figure~\ref{fig:F6}, different parameters of BF affected its performance significantly. Typically, BF generated curves with a non-linear relationship between noise level and the MAE and SSIM performance, while MKF demonstrated a near-linear correlation when $t=2$. To sum up, besides improving the noise removal capacity, MKF achieved a higher-level of adaptivity than BF.

\subsection{Adaptivity under Spatially-Varying Noise}
\label{subsec:s44}

On BrainWeb, we still employed the same parameters of BF and MKF except for the window radius, which was $2$ considering the image size became much smaller. For evaluation, we compared MKF with BF, CF, and TV. Their parameters were well chosen and listed as following. The maximal iteration of TV was $100$, and the $\lambda$ is $1.25$. We choose the Gaussian curvature in CF and set the maximal iteration $10$.

	\let\iwidth\relax
	\let\iheight\relax
	\let\rowspace\relax
	\newlength{\iwidth}
	\newlength{\iheight}
	\newlength{\rowspace}
	\setlength{\iwidth}{0.16\textwidth}
	\setlength{\iheight}{1.0\iwidth}
	\setlength{\rowspace}{0pt}		
	\newcommand\ann[1]{
		\begin{tikzpicture}[baseline, trim left]
		\node[align=center,rotate=90] at (0.05\iwidth, 0.5\iheight) {\scriptsize{#1}};
		\end{tikzpicture}
	} 
    \newcommand{\columnText}[1]{
	    \begin{tikzpicture}[baseline,trim right=3pt]
	    \node[rotate=90] at (0.05\iwidth, 0.5\iheight) {\scriptsize{#1}};
	    \end{tikzpicture}
    }
	\tikzset{
		/tpic/options/.style = {
			width=\iwidth,
			height=\iheight,
			scale=1.0,
			xloc=0.5,
			yloc=0.5,
			angle=0,
			color=white,
			label={}
		}
	}
	
	\setlength{\tabcolsep}{1.5pt}
	\begin{figure}
		\centering
		\begin{tabular}{cccc}
			\ann{Noise free} \tpic[options]{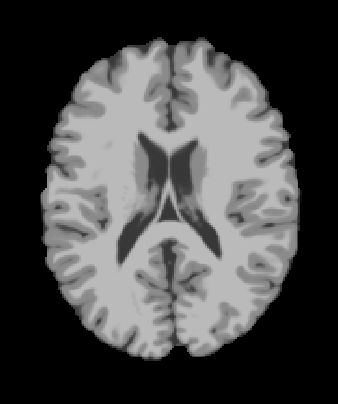} & \tpic[options]{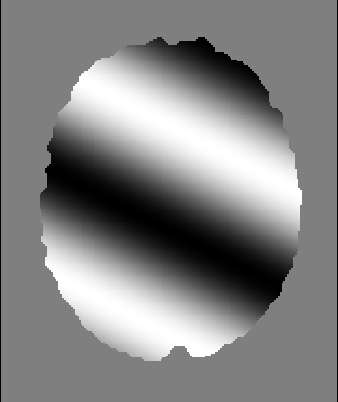} & \tpic[options]{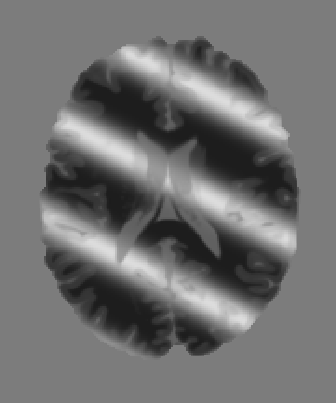} &
			\tpic[options]{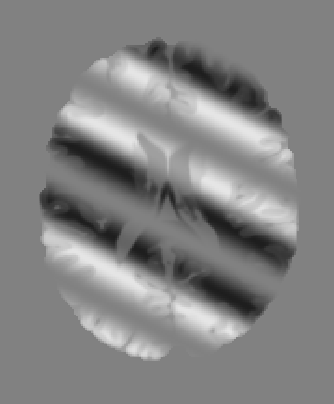} \\
			\ann{Noisy} \tpic[options]{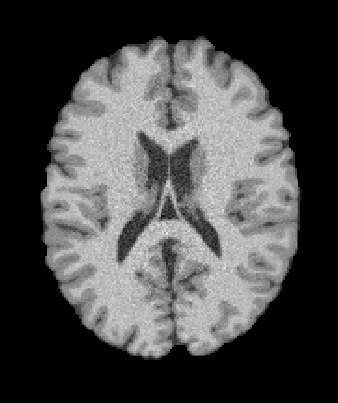} & \tpic[options]{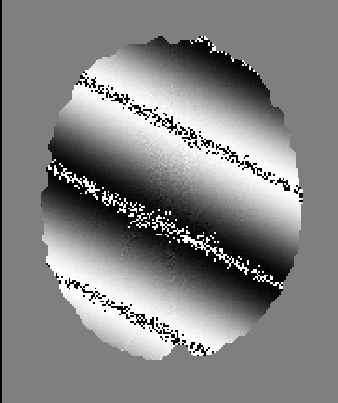} & \tpic[options]{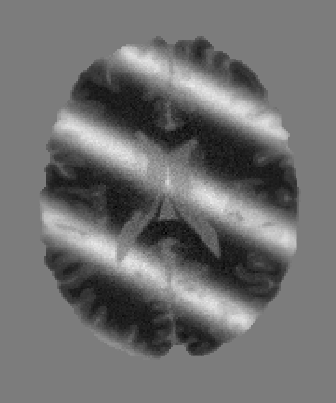} &
			\tpic[options]{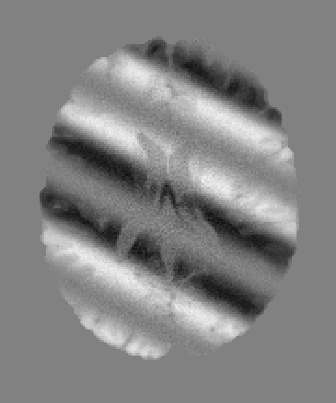} \\						
			[0.6\rowspace] \scriptsize{\ \ \ \ \ \ \ \  Magnitude} & \scriptsize{Phase} & \scriptsize{Real part} & \scriptsize{Imaginary part}
		\end{tabular}
		\caption[]{Noise free and noisy synthetic MRI magnitude, phase map, and the complex components.} 
		\label{fig:F7}
	\end{figure}

As the noise distribution of complex components of MRI data is spatially-varying Gaussian, we derived noise-free real and imaginary components using the synthesized background phase \citet{pizzolato2016noise}, and next, we added the spatially-varying noise to them.  It should point out that the phase maps are slice-by-slice varying and are gradually transitional; this typically breaks the low-rank assumption of an image. We can see the noise-free and noisy synthetic slices in Figure~\ref{fig:F7}.

	\let\iwidth\relax
	\let\iheight\relax
	\let\rowspace\relax
	\newlength{\iwidth}
	\newlength{\iheight}
	\newlength{\rowspace}
	\setlength{\iwidth}{0.14\textwidth}
	\setlength{\iheight}{1.0\iwidth}
	\setlength{\rowspace}{0pt}	 	
	\tikzset{
		/tpic/options/.style = {
			width=\iwidth,
			height=\iheight,
			scale=1.0,
			xloc=0.5,
			yloc=0.5,
			angle=0,
			color=white,
			label={}
		}
	}
    \newlength{\gbarlength}
    \setlength{\gbarlength}{1.1 \iheight}    
    \pgfdeclareverticalshading{g1}{.20cm}{ 
    	rgb(0)=(0,0,0); 
    	rgb(1.30\gbarlength)=(1,1,1)
    }
    \newcommand\gBar[2]{		
	\begin{tikzpicture} 
	\node (bar) {\pgfuseshading{g1}}; 
	\node [below=0mm of bar] {\scriptsize{#1}};
	\node [above=-0.3mm of bar] {\scriptsize{#2}};
	\end{tikzpicture}
} 	
	\setlength{\tabcolsep}{1.5pt}
\begin{figure*} 
	\centering
	\begin{tikzpicture}
	\node(dum) {
			\begin{tabular}{cccccccc}
			   \columnText{Real part} \tpic[options]{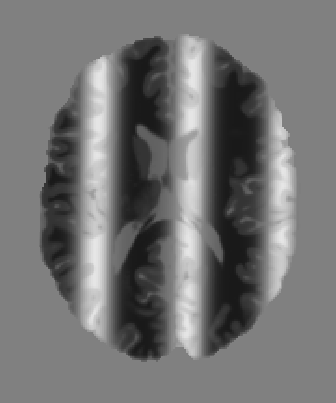}  &	 \tpic[options]{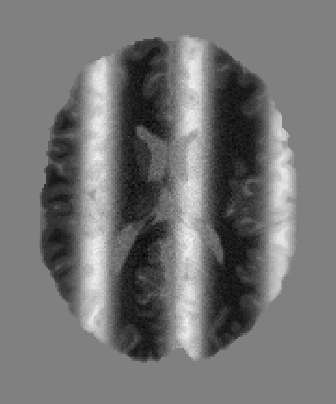} & \tpic[options]{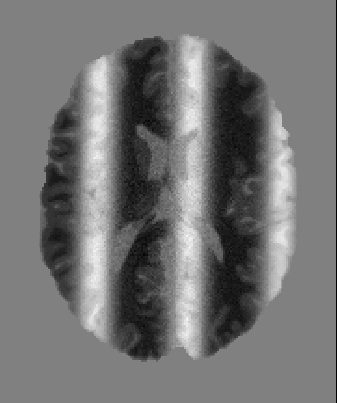} & \tpic[options]{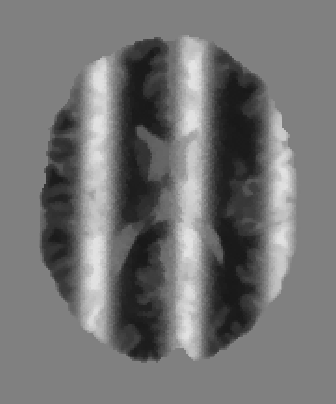} &	\tpic[options]{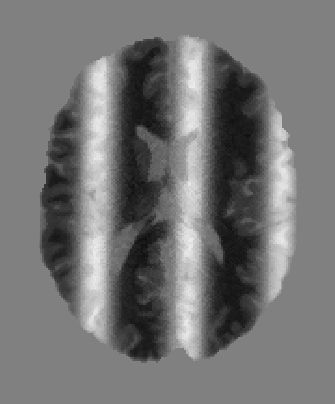} &	\tpic[options]{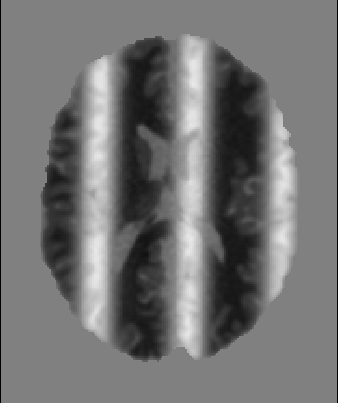}  \vspace{-0.2cm} \\	
			   \columnText{Imaginary part} \tpic[options]{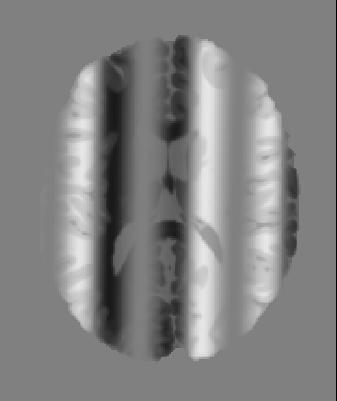} & \tpic[options]{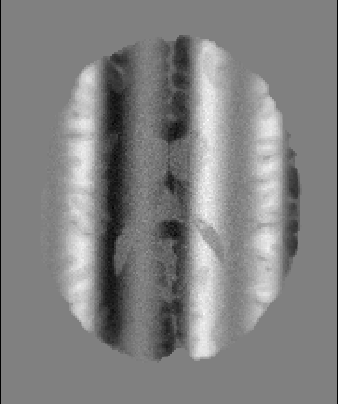} & \tpic[options]{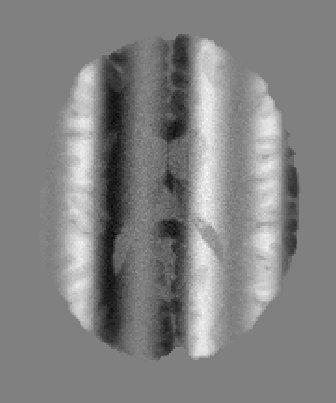} &\tpic[options]{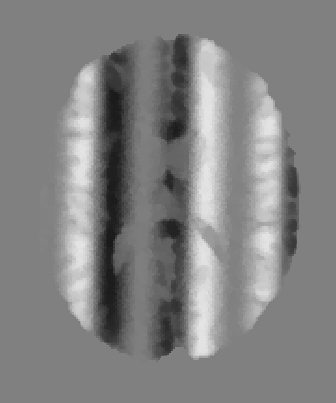} &	\tpic[options]{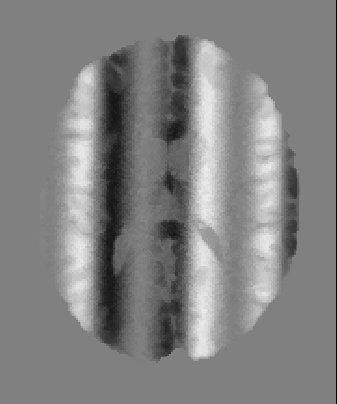} &	\tpic[options]{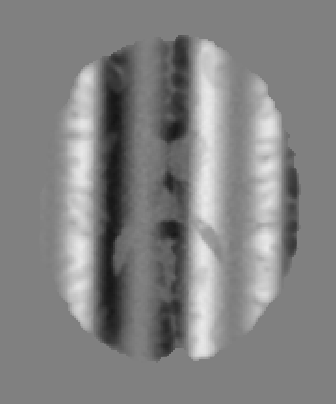}  \vspace{-0.05cm} \\	
			   [0.6\rowspace] \small{\quad \quad Noise Free}  & \small{Noisy} &  \small{BF}  & \small{TV} & \small{CF} & \small{MKF} 
	    	\end{tabular}
		};
        \node(cb) [right = -0.1cm of dum] {\gBar{$-3000$}{$3000$}};
		\end{tikzpicture}  
		\caption{Both real and imaginary images demonstrate promising denoising performance of MKF.}  
		\label{fig:F8}
\end{figure*}

Figure~\ref{fig:F8} shows noise-free (first column) and noisy slices (second column) of both real and imaginary parts, as well as the denoised images from $4$ filters.  We can see in Figure~\ref{fig:F8}, the noise severely corrupts detailed image contents. After denoising, however, BF still reserved a lot of noise points. Although TV and CF achieved better results, they demonstrated mosaic blocks; and MKF generated smoother results which were closer to the noise-free ground truth. For quantitative evaluation, we further generated the MAE and SSIM curves.   



  \begin{figure}
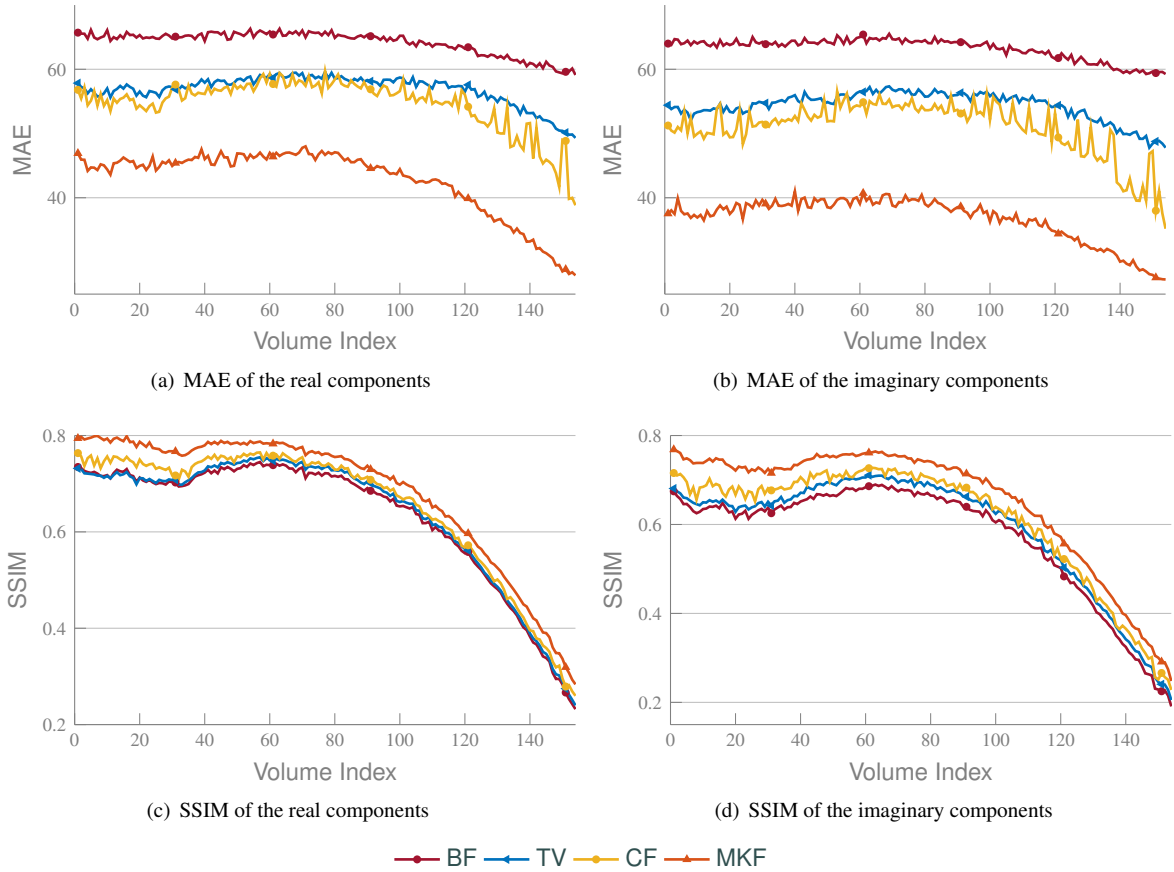
          	
	\centering
	\subfigure[MAE of the real components]{
		\label{fig:sub:f10a}
		\subimport{/}{mae_re.tex} 
	}
	\subfigure[MAE of the imaginary components]{
		\label{fig:sub:f10b}
		\subimport{/}{mae_im.tex} 
	}
	\\
	\subfigure[SSIM of the real components]{
		\label{fig:sub:f10c}
		\subimport{/}{ssim_re.tex}
	}
	\subfigure[SSIM of the imaginary components]{
		\label{fig:sub:f10d}
		\subimport{/}{ssim_im.tex}
	}
	\\
    \definecolor{mycolor1}{rgb}{0.00000,0.44700,0.74100}%
    \definecolor{mycolor2}{rgb}{0.46600,0.67400,0.18800}%
    \definecolor{mycolor3}{rgb}{0.85000,0.32500,0.09800}%
    \definecolor{mycolor4}{rgb}{0.30100,0.74500,0.93300}%
    \definecolor{mycolor5}{rgb}{0.92900,0.69400,0.12500}%
    \definecolor{mycolor6}{rgb}{0.63500,0.07800,0.18400}%
	\ref{legend:M_S4}
	\label{fig:F10}       
	\caption{MKF achieves better MAE and SSIM indices than BF, TV, and CF.}
\end{figure}

As shown in Figure~\ref{fig:sub:f10a}, \ref{fig:sub:f10b}, \ref{fig:sub:f10c}, and  \ref{fig:sub:f10d}, MAE and SSIM curves illustrate the quantitative performance. The curves demonstrate the MAE and SSIM values derived from synthetic data. MKF achieved a better MAE and SSIM performance than BF, TV, and CF. Although BF achieved competitive performance  shown in Figure~\ref{fig:F6}, after the image characteristics have been changed, the same parameters degenerate its performance significantly. By contrast, with the same set of parameters, MKF still performed robustly demonstrating its high-level adaptivity. 

\section{Conclusion}
\label{sec:s5}

To improve filtering adaptivity, we extended bilateral filtering using image context and proposed multi-kernel filter. The image context was heuristically constructed through the proposed clustering method, which followed Gestalt grouping rules. As a result, the range kernel can be automatically initialized by the image content itself. With this extension, MKF demonstrates a higher-level adaptivity than BF and shows promising noise-removal performance outperforming three state-of-the-art filters, including BF, TV, and CF. As far as we know, this is the first study investigating and simulating the adaptive visual mechanisms in designing a smoothing filter. However, some limitations are worth noting. Although MKF adaptively generated parameters of filtering kernels, its performance depended on the parameters of the clustering. In addition, MKF was relatively low in time efficiency owing to the hierarchically clustering. Future works should develop a more effective computational model of vision adaptivity for addressing such limitations. 
	
	
	\bibliographystyle{cas-model2-names}
	
	\bibliography{cas-refs}

\end{document}